\documentclass[10pt,twocolumn,letterpaper]{article}

\usepackage{iccv}
\usepackage{times}
\usepackage{epsfig}
\usepackage{graphicx}
\usepackage{amsmath,bm}
\usepackage{amssymb}
\usepackage{nicefrac}

\usepackage{bm}
\newcommand{\mathbold}[1]{\bm{#1}}
\newcommand{\mbf}[1]{\mathbf{#1}}

\newcommand{\vectb}[1]{\mathbold{#1}}

\newcommand{\T}{\top}

\DeclareMathOperator{\diag}{diag}

\newcommand{\vmu}[0]{\mathbold{\mu}}

\newcommand{\MPhi}[0]{\mathbold{\Phi}}
\newcommand{\MSigma}[0]{\mathbold{\Sigma}}

\renewcommand{\mid}[0]{\,|\,}

\newcommand{\vh}{\mbf{h}}

\newcommand{\vk}{\mbf{k}}

\newcommand{\vt}{\mbf{t}}

\newcommand{\vy}{\mbf{y}}
\newcommand{\vz}{\mbf{z}}

\newcommand{\MC}{\mbf{C}}

\newcommand{\MH}{\mbf{H}}
\newcommand{\MI}{\mbf{I}}

\newcommand{\MK}{\mbf{K}}

\newcommand{\MQ}{\mbf{Q}}
\newcommand{\MR}{\mbf{R}}

\newcommand{\MY}{\mbf{Y}}
\newcommand{\MZ}{\mbf{Z}}

\definecolor{mycolor0}{rgb}{0.2667,0.4471,0.7098}
\definecolor{mycolor1}{rgb}{0.1647,0.6706,0.3804}
\definecolor{mycolor2}{rgb}{0.8275,0.2627,0.3059}
\definecolor{mycolor3}{rgb}{0.5216,0.4392,0.7176}
\definecolor{mycolor4}{rgb}{0.8118,0.7255,0.4118}
\definecolor{mycolor5}{rgb}{0.2745,0.7176,0.8157}

\definecolor{mylcolor0}{rgb}{0.6902,0.7686,0.8863}
\definecolor{mylcolor1}{rgb}{0.5451,0.8902,0.6941}
\definecolor{mylcolor2}{rgb}{0.9412,0.7490,0.7647}
\definecolor{mylcolor3}{rgb}{0.8627,0.8392,0.9176}
\definecolor{mylcolor4}{rgb}{0.9569,0.9373,0.8667}
\definecolor{mylcolor5}{rgb}{0.7529,0.9020,0.9373}
\definecolor{mylcolor6}{rgb}{0.8750,0.8750,0.8750}

\usepackage{tikz,pgfplots}
\usetikzlibrary{plotmarks,shapes,arrows,positioning,3d}
\usetikzlibrary{arrows.meta}
\usetikzlibrary{decorations.text}
\pgfplotsset{compat=newest} 

\usetikzlibrary{quotes,arrows.meta,decorations.markings}
\def\edgecolor{rgb:blue,4;red,1;green,4;black,3}
\newcommand{\midarrow}{\tikz \draw[-Stealth,line width =0.8mm,draw=\edgecolor] (-0.3,0) -- ++(0.3,0);}
\usepackage{Ball}
\usepackage{Box}
\usepackage{RightBandedBox}

\newlength\figureheight
\newlength\figurewidth

\usepackage{booktabs}
\usepackage{array} 
\usepackage{multirow}

\usepackage[font=small,labelfont=small,skip=4pt,labelsep=period]{caption}
\usepackage{subcaption}

\newcolumntype{P}[1]{>{\centering\arraybackslash}p{#1}}

\renewcommand{\paragraph}[1]{~\\[-3pt]\noindent\textbf{#1.}~~}

\usepackage[pagebackref=true,breaklinks=true,letterpaper=true,colorlinks,bookmarks=false]{hyperref}
\iccvfinalcopy

\def\iccvPaperID{6035}

\ificcvfinal\fi
\begin{document}

\title{Multi-View Stereo by Temporal Nonparametric Fusion}

\author{Yuxin Hou \qquad\quad Juho Kannala \qquad\quad Arno Solin \\
Department of Computer Science\\
Aalto University, Finland \\
{\tt\small firstname.lastname@aalto.fi}
}

\maketitle
\ificcvfinal\thispagestyle{empty}\fi

\begin{abstract}
  We propose a novel idea for depth estimation from multi-view image-pose pairs, where the model has capability to leverage information from previous latent-space encodings of the scene. This model uses pairs of images and poses, which are passed through an encoder--decoder model for disparity estimation. The novelty lies in soft-constraining the bottleneck layer by a nonparametric Gaussian process prior. We propose a pose-kernel structure that encourages similar poses to have resembling latent spaces. The flexibility of the Gaussian process (GP) prior provides adapting memory for fusing information from previous views. We train the encoder--decoder and the GP hyperparameters jointly end-to-end. In addition to a batch method, we derive a lightweight estimation scheme that circumvents standard pitfalls in scaling Gaussian process inference, and demonstrate how our scheme can run in real-time on smart devices.
\end{abstract}

\section{Introduction}
\label{sec:intro}
\noindent
Multi-view stereo (MVS) refers to the problem of reconstructing 3D scene structure from multiple images with known camera poses and internal parameters. For example, estimation of depth maps from multiple video frames captured by a moving monocular video camera \cite{wang2018mvdepthnet} is a variant of MVS when the motion is known. Other variants of the problem include depth estimation using conventional two-view stereo rigs \cite{kendall} and image-based 3D modelling from image collections \cite{Furukawa_internetscalemvs,Schonberger+Zheng+Frahm+Pollefeys:2016}. MVS reconstructions have various applications.
For instance, image-based 3D models can be used for measurement and visualization of large environments to aid design and planning \cite{acute3d}, and depth estimation from stereo rigs or monocular videos benefits perception and simultaneous localization and mapping (SLAM) in the context of autonomous machines.

In this paper, we focus on depth map estimation for video frames captured by a monocular camera, whose motion is unconstrained but known. In practice, the motion could be estimated using visual-inertial odometry techniques (see, \eg, \cite{solin2018pivo}), which are capable of providing high-precision camera poses in real-time with very small drift and are also commonly available in standard mobile platforms (\eg,~ARCore on Android and ARKit on iOS).

\begin{figure}[!t]
  \small
  \setlength{\figurewidth}{.25\columnwidth}
  \setlength{\figureheight}{0.75\figurewidth}    
  
  \begin{subfigure}{\columnwidth}\centering
  \begin{tikzpicture}

    \begin{scope}
      \clip[rounded corners=2mm] (0,0) rectangle coordinate (centerpoint) ({2*\figurewidth},\figureheight); 
      \node[inner sep=0pt] at (.5\figurewidth,.5\figureheight) {\includegraphics[width=\figurewidth]{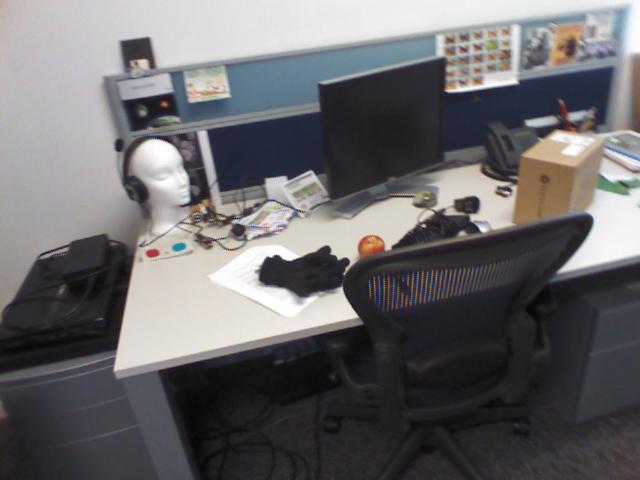}}; 
    \end{scope}

    \begin{scope}
      \clip[rounded corners=2mm] (2*\figurewidth,0) rectangle coordinate (centerpoint) ({3.03*\figurewidth+\figurewidth},\figureheight); 
      \node[inner sep=0pt] at (.5\figurewidth+3.03*\figurewidth,.5\figureheight) {\includegraphics[width=\figurewidth]{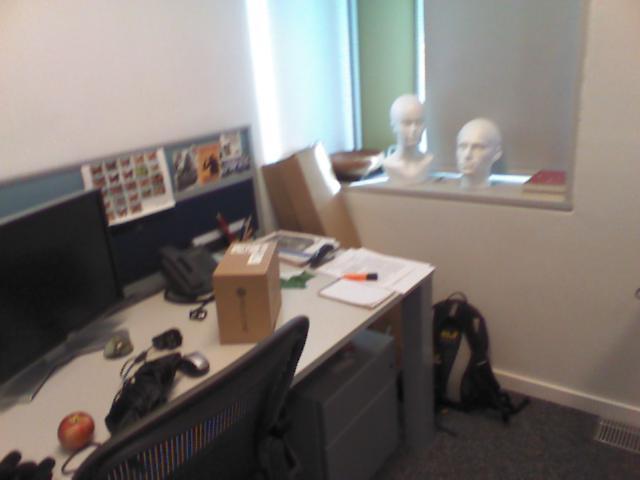}}; 
    \end{scope}

    \foreach \x in {1,...,2} {
      \node (\x) at ({\x*\figurewidth+.5*\figurewidth+0.01*\figurewidth*\x},.5*\figureheight) {\includegraphics[width=\figurewidth]{fig/figure1/00\x-ref.jpg}};  
    }
  \end{tikzpicture}
  \caption{Reference frames}
  \end{subfigure}
  \begin{subfigure}{\columnwidth}\centering
  \begin{tikzpicture}

    \begin{scope}
      \clip[rounded corners=2mm] (0,0) rectangle coordinate (centerpoint) ({2*\figurewidth},\figureheight); 
      \node[inner sep=0pt] at (.5\figurewidth,.5\figureheight) {\includegraphics[width=\figurewidth]{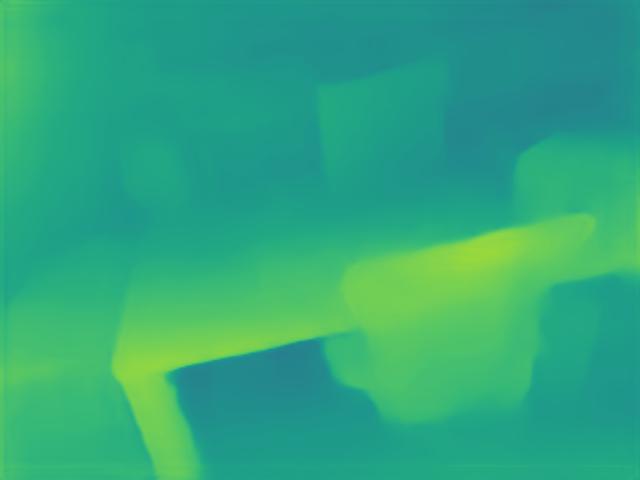}}; 
    \end{scope}

    \begin{scope}
      \clip[rounded corners=2mm] (2*\figurewidth,0) rectangle coordinate (centerpoint) ({3.03*\figurewidth+\figurewidth},\figureheight); 
      \node[inner sep=0pt] at (.5\figurewidth+3.03*\figurewidth,.5\figureheight) {\includegraphics[width=\figurewidth]{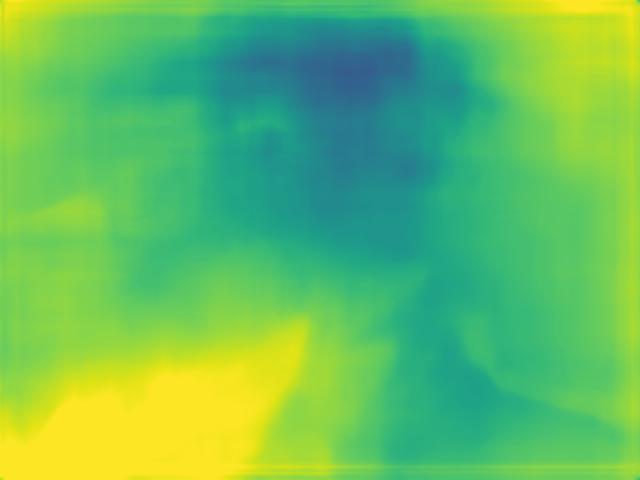}}; 
    \end{scope}

    \foreach \x in {1,...,2} {
      \node (\x) at ({\x*\figurewidth+.5*\figurewidth+0.01*\figurewidth*\x},.5*\figureheight) {\includegraphics[width=\figurewidth]{fig/figure1/00\x-wo}};
            
   }
  \end{tikzpicture}
  \caption{Multi-view depth-estimation w/o GP}
  \end{subfigure}
  \begin{subfigure}{\columnwidth}\centering
  \begin{tikzpicture}

    \begin{scope}
      \clip[rounded corners=2mm] (0,0) rectangle coordinate (centerpoint) ({2*\figurewidth},\figureheight); 
      \node[inner sep=0pt] at (.5\figurewidth,.5\figureheight) {\includegraphics[width=\figurewidth]{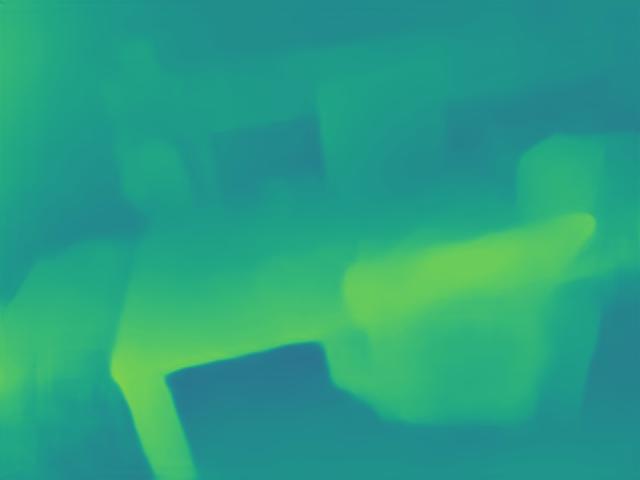}}; 
    \end{scope}

    \begin{scope}
      \clip[rounded corners=2mm] (2*\figurewidth,0) rectangle coordinate (centerpoint) ({3.03*\figurewidth+\figurewidth},\figureheight); 
      \node[inner sep=0pt] at (.5\figurewidth+3.03*\figurewidth,.5\figureheight) {\includegraphics[width=\figurewidth]{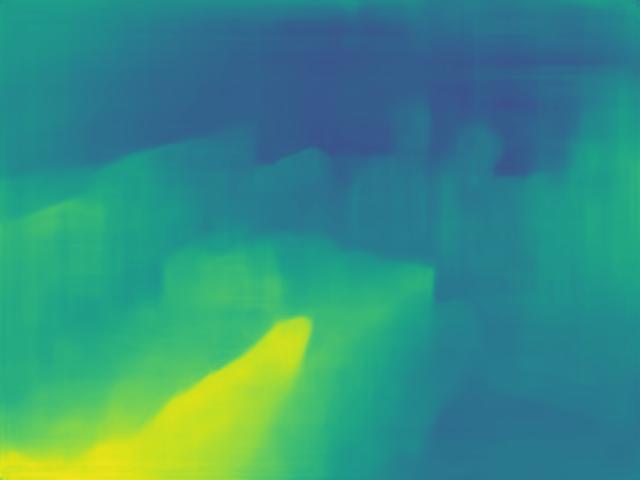}}; 
    \end{scope}

    \foreach \x in {1,...,2} {
      \node (\x) at ({\x*\figurewidth+.5*\figurewidth+0.01*\figurewidth*\x},.5*\figureheight) {\includegraphics[width=\figurewidth]{fig/figure1/00\x-w.jpg}};
    }
  \end{tikzpicture}
  \caption{Multi-view depth-estimation with GP}
  \end{subfigure}

  \caption{An example sequence of depth estimation results, where introducing information sharing in the latent space helps improving the depth maps by making them more stable and edges sharper.}
  \label{fig:teaser}
\end{figure}

Depth estimation from multiple video frames under varying and arbitrary motion is more challenging than depth estimation using a rigid two-view stereo rig, but there can be potential benefits in using a moving monocular camera instead of a fixed rig. Firstly, in small mobile devices the baseline between the two cameras of the rig can not be large and this limits the range of depth measurements. With a moving monocular camera the motion usually provides a larger baseline than the size of the device and thus measurement accuracy for distant regions can be improved. Secondly, when the camera is translating and rotating in a given space, it typically observes the same scene regions from multiple continuously varying viewpoints, and it would be beneficial to be able to effectively fuse all this information for more robust and stable depth estimation. 

In this work, we propose a new approach that combines a disparity estimation network, which has an encoder--decoder architecture and plane-sweep cost volume input as in \cite{wang2018mvdepthnet}, and a Gaussian process (GP, \cite{Rasmussen+Williams:2006}) prior, which soft-constrains the bottleneck layer of the network to fuse information from video frames having similar poses. This is achieved by proposing a pose-kernel structure which encourages similar poses to have resembling latent space representations. The motivation behind the proposed approach is to efficiently improve fusion of information from overlapping views independently of their separation in time. That is, our pose-kernel can implicitly fuse information from all frames, which have overlapping fields of view, and without making the prediction of individual depth maps more time-consuming or without spending additional effort in the cost volume computation. In contrast to hard and heuristic view selection rules that are often applied in similar context \cite{wang2018mvdepthnet,yao2018mvsnet} our approach allows soft fusion of information via the latent representation. Our approach can be applied either in batch mode, where the fused result utilizes all available frames, or in online mode, where only previous frames affect the prediction of the current frame.

The contributions of this paper are as follows. {\em (i)}~We propose a novel approach for multi-view stereo that passes information from previously reconstructed depth maps through a probabilistic prior in the latent space; {\em (ii)}~For the non-parametric latent-space prior we propose a pose-kernel approach for encoding prior knowledge about effects of the relative camera pose between observation frames; {\em (iii)}~We show that the CNN encoder--decoder structure and the GP hyperparameters can be trained jointly; {\em (iv)}~We extend our method to an online scheme capable of running in real-time in smartphones/tablets.

To our knowledge this is the first paper to utilize GP priors for multi-view information fusion and also the first attempt at scalable online MVS on smartdevices.

\section{Related work}
\label{sec:related}
\noindent
MVS approaches can be categorized based on their output representations as follows: \emph{(a)}~volumetric reconstruction methods \cite{Kutulakos, Kolev,kar2017learning}, \emph{(b)}~point cloud reconstruction methods \cite{Furukawa_pmvs, Ylimaki}, and \emph{(c)}~depth map based methods \cite{yao2018mvsnet}. In many cases, point cloud and depth map representations are finally converted to a triangular surface mesh for refinement \cite{Furukawa_pmvs,keriven}. Volumetric voxel based approaches have shown good performance with small objects but are difficult to apply for large scenes due to their high memory load. Point cloud based approaches provide accurate reconstructions for textured scenes and objects but scenes with textureless surfaces and repeating patterns are challenging. In this work, we focus on multi-view depth estimation since depth map based approaches are flexible and suitable for most use cases.

There has recently been plenty of progress in learning-based depth estimation approaches. Inspired by classical MVS methods \cite{collins}, most attempts on learned MVS use plane-sweeping approaches to first compute a matching cost volume from nearby images and then regard depth estimation as a regression or multi-class classification problem, which is addressed by deep neural networks \cite{huang2018deepmvs,wang2018mvdepthnet,yao2018mvsnet}. DeepTAM~\cite{zhou2018deeptam} computes the sum of absolute difference of patches between warped image pairs and use an adaptive narrow band strategy to increase the density of sampled planes. DeepMVS~\cite{huang2018deepmvs} proposed a patch matching network to extract features to aid in the comparison of patches. For feature aggregation, it considers both an intra-volume feature aggregation network and inter-volume aggregation network. MVDepthNet~\cite{wang2018mvdepthnet} computes the absolute difference directly without a supporting window to generate the cost volume, as the pixel-wise cost matching enable the volume to preserve detail information. MVSNet~\cite{yao2018mvsnet} proposes a variance-based cost metric and employ a 3D CNN to obtain a smooth cost volume automatically. DPSNet~\cite{im2018dpsnet} concatenate warped features and use a series of 3D convolutions to learn the cost volume generation.

It is important to note that none of the aforementioned learning based MVS approaches has been demonstrated on a mobile platform. Indeed, most of the methods are heavy and it takes several seconds or even more to evaluate a single depth map with a powerful desktop GPU \cite{huang2018deepmvs,yao2018mvsnet}. The most light-weight model is \cite{wang2018mvdepthnet}, and therefore we use it as a baseline upon which we add our complementary contributions. The monocular depth estimation system in \cite{wang2018mvdepthnet} uses a view selection rule, which selects frames that have enough angle or translation difference and then uses the selected frames for computing the cost volume. However, this kind of view selection can not use information from similar views in the more distant past, since the future motion is unknown and all past frames can not be stored. In contrast, our approach allows to utilize all past information in a computationally efficient manner. Also, our contribution is not competing with the various network architectures proposed recently \cite{huang2018deepmvs,zhou2018deeptam,yao2018mvsnet,wang2018mvdepthnet,hou2019unstructured} but complementary: The temporal coupling of latent representations has not been proposed earlier and could be combined also with other network architectures than \cite{wang2018mvdepthnet}, which we use in our experiments.

Another area of related work is depth map fusion which aims to integrate multiple depth maps into a unified scene representation and needs to deal with inconsistencies and redundancies in the process.  For example, \cite{merrell2007real} defines three types of visibility relationships between predicted depth maps, determining the validity of estimation by detecting occlusions and free-space violations. Also, volumetric approaches are widely used for fusion and reconstruction \cite{newcombe2011kinectfusion, niessner2013real}. Again, our method is complementary: it shares information implicitly in the latent space, and can be combined with a depth map fusion post-processing stage.

Finally, regarding the technical and methodological aspects of our work, we combine both deep neural networks and Gaussian process (GP) models. GPs are a probabilistic machine learning paradigm for encoding flexible priors over functions \cite{Rasmussen+Williams:2006}. They have not been much used in this area of geometric computer vision. Though, GPs have been used in other latent variable modelling tasks in vision, where uncertainty quantification \cite{Kendall+Gal:2017} plays a crucial role---including variational autoencoders with GP priors \cite{eleftheriadis2016variational,Casale+Dalca+Saglietti+Listgarten+Fusi:2018} and GP based latent variable models for multi-view and view-invariant facial expression recognition \cite{eleftheriadis2015discriminative,eleftheriadis2015multi}. In \cite{Casale+Dalca+Saglietti+Listgarten+Fusi:2018} GPs are applied to face image modelling, where the GP kernel accounts for the pose, and in \cite{Urtasun:2006} they are used for 3D people tracking. The motivation for our work is in recent advances in real-time inference using GPs \cite{Sarkka+Solin+Hartikainen:2013,Solin+Hensman+Turner:2018} that make them applicable to online inference in smartphones.

\section{Methods}
\label{sec:methods}
\begin{figure}
  \centering
  \resizebox{1.03\columnwidth}{!}{%
  \footnotesize 
  \begin{tikzpicture}

    \newcommand{\drawnet}[3]{%
      \tikzstyle{block} = [rounded corners=0.5pt,minimum width=1mm,minimum height=1mm,inner sep=0,draw=#3!50!black,fill=#3]
      \foreach \j in {0,...,4} 
        \node[block,minimum height={0.25cm+0.2cm*\j}] at (#1-.1*\j,#2) {};  
      \foreach \j in {0,...,4} 
        \node[block,minimum height={0.25cm+0.2cm*\j}] at (#1+.1*\j,#2) {};}

    \newcommand{\encoder}[3]{%
      \tikzstyle{block} = [rounded corners=0.5pt,minimum width=1mm,minimum height=1mm,inner sep=0,draw=#3!50!black,fill=#3]
      \foreach \j in {0,...,4} 
        \node[block,minimum width={0.25cm+0.2cm*\j}] at (#1,#2+.1*\j) {};} 

    \newcommand{\decoder}[3]{%
      \tikzstyle{block} = [rounded corners=0.5pt,minimum width=1mm,minimum height=1mm,inner sep=0,draw=#3!50!black,fill=#3]
      \foreach \j in {0,...,4} 
        \node[block,minimum width={0.25cm+0.2cm*\j}] at (#1,#2-.1*\j) {};} 

    \newcommand{\costvol}[2]{%
      \foreach \y in {1,...,6} {
        \draw[draw=mycolor1,fill=mycolor1!10,rounded corners=1pt] ({-0.75+0.05*\y+2*#1}, {#2+.07*\y}) -- ++({1.5-.1*\y},0) -- ++({-.25+.03*\y},{0.5-0.05*\y}) -- ++({-1+0.05*\y},0) -- cycle;}}

    \tikzstyle{arrow} = [draw=black!10, single arrow, minimum height=10mm, minimum width=3mm, single arrow head extend=1mm, fill=black!10, anchor=center, rotate=-90, inner sep=2pt]

    \node[rotate=90] at (-1,6.5) {Camera pose};    
    \node[rotate=90] at (-1,4.5) {Frame};
    \node[rotate=90] at (-1,3.25) {Cost vol.};
    \node[rotate=90] at (-1,2.0) {Encoder};
    \node[rotate=90] at (-1,0) {Latent GP};
    \node[rotate=90] at (-1,-1.25) {Decoder};
    \node[rotate=90] at (-1,-2.5) {Disparity};

     \draw [dashed,mycolor0] (0,6.5) to [out=-60,in=90] (1,4.5);
     \draw [dashed,mycolor0] (2,6) to [out=-140,in=90] (1,4.5);

     \draw [dashed,mycolor0] (2,6) to [out=-40,in=90] (3,4.5);
     \draw [dashed,mycolor0] (4,6.3) to [out=-140,in=90] (3,4.5);
     
     \draw [dashed,mycolor0] (4,6.3) to [out=-40,in=90] (5,4.5);
     \draw [dashed,mycolor0] (6,7) to [out=-140,in=90] (5,4.5);

     \draw [dashed,mycolor0] (6,7) to [out=-40,in=90] (7,4.5);
     \draw [dashed,mycolor0] (8,6.8) to [out=-140,in=90] (7,4.5);
     
     \node[rotate=90] at (0.8,2) {\textcolor{mycolor0}{\scriptsize Pose similarity}};

     \draw[thick,mycolor0,-{Triangle[angle=60:3pt 2]},postaction={decorate,decoration={raise=-2.5ex,text along path,text align=center,text={||~~~~~~~~~~~~~~~~~~~~~~~~~~~~~~~~~~~~~~~~~~~~~~~~~~Camera pose trajectory}}}] plot[smooth,tension=1]
       coordinates{(0,6.5) (2,6) (4,6.3) (6,7) (8,6.8) (9,6.5)};
       
     \draw[mycolor0] plot coordinates{(0.0544,7.2813) (0.7839,6.5975) (0.6061,6.4285) (-0.1234,7.1123) (0.0544,7.2813) (0.0000,6.5000) (0.7839,6.5975) (0.0000,6.5000) (-0.1234,7.1123) (0.0000,6.5000) (0.6061,6.4285) };
     
     \draw[mycolor0] plot coordinates{(1.6408,6.7356) (2.6207,6.5361) (2.5430,6.1756) (1.5631,6.3751) (1.6408,6.7356) (2.0000,6.0000) (2.6207,6.5361) (2.0000,6.0000) (1.5631,6.3751) (2.0000,6.0000) (2.5430,6.1756) };

     \draw[mycolor0] plot coordinates{(4.0913,7.1167) (4.8136,6.4252) (4.5467,6.1520) (3.8243,6.8435) (4.0913,7.1167) (4.0000,6.3000) (4.8136,6.4252) (4.0000,6.3000) (3.8243,6.8435) (4.0000,6.3000) (4.5467,6.1520) };
 
     \draw[mycolor0] plot coordinates{(5.5133,7.6429) (6.5131,7.6302) (6.4932,7.3015) (5.4934,7.3142) (5.5133,7.6429) (6.0000,7.0000) (6.5131,7.6302) (6.0000,7.0000) (5.4934,7.3142) (6.0000,7.0000) (6.4932,7.3015) };
      
     \draw[mycolor0] plot coordinates{(7.4043,7.2899) (8.3542,7.5561) (8.2898,7.2097) (7.3399,6.9434) (7.4043,7.2899) (8.0000,6.8000) (8.3542,7.5561) (8.0000,6.8000) (7.3399,6.9434) (8.0000,6.8000) (8.2898,7.2097) };

     \node[shape=rectangle,draw=black!50,minimum width=1.5cm,minimum height=1cm, rounded corners=1pt, inner sep=1pt] (I0) at (0,4.5) {\includegraphics[width=1.5cm]{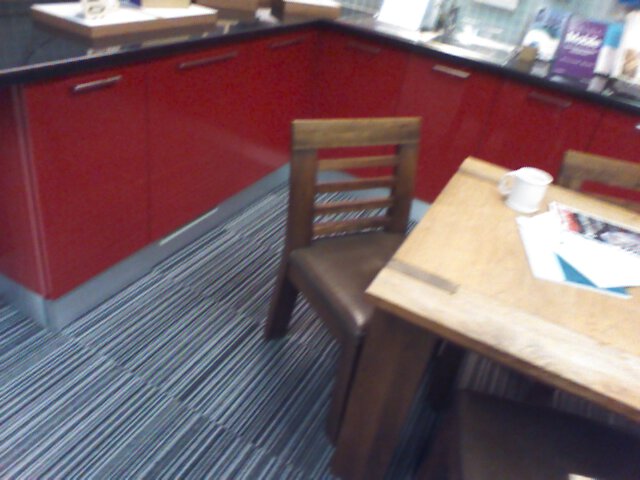}};

    \node[shape=circle,draw=mycolor0,fill=mycolor0!50,thick,minimum width=2em] (z0) at (0,0) {$\vz_0$};

    \foreach \x in {1,...,4} {

      \node[shape=rectangle,draw=black!50,minimum width=1.5cm,minimum height=1cm, rounded corners=1pt, inner sep=1pt] (I\x) at (2*\x,4.5) {\includegraphics[width=1.5cm]{fig/sequence/00\x-ref.jpg}};

      \draw[thick, draw=mycolor1,-latex] ({2*\x-1.5},{3.9}) to [bend left=-35] ({2*\x-.6},{3.25});
      \draw[thick, draw=mycolor1,-latex] ({2*\x},{3.9}) -- ({2*\x},{2.5});

      \costvol{\x}{2.8}

      \draw[-latex,dashed] ({2*\x-0.4},2.4) to [bend right=20] node[below,rotate=90] {\scriptsize skip} ++(0,-3.6);

      \encoder{{2*\x}}{2}{mycolor4}
      \node (e\x) at ({2*\x},2) {};

      \node[shape=circle,draw=mycolor2,fill=mycolor2!50,thick,minimum width=1.5em] (y\x) at ({2*\x},1) {\tiny $\vy_\x$};

      \draw[-latex,thick,mycolor2] (e\x)->(y\x);

      \node[shape=circle,draw=mycolor0,fill=mycolor0!50,thick,minimum width=2em] (z\x) at ({2*\x},0) {$\vz_\x$};

      \draw[-latex,thick,mycolor2] (y\x)->(z\x);

      \decoder{{2*\x}}{-1}{mycolor4}
      \node (d\x) at ({2*\x},-1) {};

      \draw[-latex,thick,mycolor0] (z\x)->(d\x);

      \node[shape=rectangle,draw=black!50,minimum width=1.5cm,minimum height=1cm, rounded corners=1pt,inner sep=1pt] (disp\x) at (2*\x,-2.5) {\includegraphics[width=1.5cm]{fig/sequence/00\x-ours.jpg}};

      \draw[-latex] (d\x)++(0,-.5)->(disp\x);

      \draw[-latex,dashed,mycolor0] ({2*\x-1},4.5) -- ({2*\x-1},0);
    
    }

    \node (z5) at (9,0) {};

    \draw[-latex,draw=mycolor0,thick] (z0)->(z1);
    \draw[-latex,draw=mycolor0,thick] (z1)->(z2);
    \draw[-latex,draw=mycolor0,thick] (z2)->(z3);
    \draw[-latex,draw=mycolor0,thick] (z3)->(z4);
    \draw[-latex,draw=mycolor0,thick] (z4)->(z5);
  
  \end{tikzpicture}}

  \caption{Illustrative sketch of our MVS approach. The camera poses and input frames are illustrated in the top rows. The current and previous (or a sequence of previous) frames are used for composing a cost volume, which is then passed through and encoder network. The novelty in our method is in doing Gaussian process inference on the latent-space encodings such that the GP prior is defined to be smooth in pose-difference. The GP prediction is finally passed through a decoder network which outputs disparity maps (bottom). This is the logic of the online variant of our method (the latent space graph is a directed graph / Markov chain). The batch variant could be illustrated in similar fashion, but with links between all latent nodes $\vz_i$.}
  \label{fig:architecture}
\end{figure}

\noindent
Our multi-view stereo approach consist of two orthogonal parts. The first (vertical data flow in Fig.~\ref{fig:architecture}) is an CNN-powered MVS approach where the input frames are warped into a cost volume and then passed through an encoder--decoder model to produce the disparity (reciprocal of depth) map. The second part (horizontal data flow in Fig.~\ref{fig:architecture}) is for coupling each of the independent disparity prediction tasks, by passing information about the latent space (bottleneck layer encodings) over the camera trajectory. We will first go through the setup in the former (Sec.~\ref{sec:mvs}), and then focus on the latter (Secs.~\ref{sec:pose-kernel}--\ref{sec:gp-online}).

\subsection{Network architecture}
\label{sec:mvs}
\noindent
For the encoder and decoder, we build upon the straightforward model in  \cite{wang2018mvdepthnet}. Our framework only includes one encoder--decoder without change of architecture, so we can compare the results directly to check the impacts of Gaussian process prior. The output of the encoder--decoder is the continuous inverse depth (disparity) prediction.  For each image-pose pair, we compute a cost volume of size $D {\times} H {\times} W$ and concatenate the reference RGB image as the input for the encoder. In this paper, we use an image size of $320{\times}256$, and $D=64$ depth planes uniformly sampled in inverse depth from $0.5$~m to $50$~m. To compute the cost volume, we warp the neighbour frame via the fronto-parallel planes at fixed depths to the reference frame using the planar homography:
\begin{equation}
  \MH = \MK \left(\MR + \vt \begin{pmatrix} 0 & 0 & \frac{1}{d_i} \end{pmatrix}\right)\MK^{-1},
\end{equation}
where $\MK$ is the known intrinsics matrix and the relative pose $(\mathbf{R}, \mathbf{t})$ is given in terms of a rotation matrix and translation vector with respect to the neighbour frame. $d_i$ denotes the depth value of the $i$\textsuperscript{th} virtual plane. The absolute intensity difference between the warped neighbour frame and the reference frame is calculated as the cost for each pixel at every depth plane:
  $V(d_i) = \sum_{R,G,B} \tilde{I}_{d_i} - I_\mathrm{r},$
where $\tilde{I}_{d_i}$ denotes the warped image via the depth plane at $d_i$ and $I_\mathrm{r}$ denotes the reference frame.

\begin{figure*}[!t]
  \centering

  \begin{subfigure}[b]{.33\textwidth}
    \centering
    \begin{minipage}{\textwidth}\tiny
    \setlength{\figurewidth}{\textwidth}
    \setlength{\figureheight}{0.2635\figurewidth}  
    \input{./fig/posekernel-track.tex}
    \end{minipage}    
    \\
    \begin{minipage}{\textwidth}
    \setlength{\figurewidth}{.25\textwidth}
    \setlength{\figureheight}{.75\figurewidth}
    \begin{tikzpicture}[inner sep=0]
    \foreach \x [count=\i] in {1,...,12}
      \node[draw=white,inner sep=0pt]
        (\i) at ({\figurewidth*mod(\i-1,4)},{-\figureheight*int((\i-1)/4)})
        {\includegraphics[width=\figurewidth]{./fig/posekernel-frame-\x.jpg}}; 
     \foreach \x [count=\i] in {1,...,12}
      \node[inner sep=0pt]
        (\i) at ({\figurewidth*mod(\i-1,4)-.4\figurewidth},{-\figureheight*int((\i-1)/4)-.4\figureheight})
        {\scriptsize\textcolor{white}{\bf\x}}; 
    \end{tikzpicture}
    \end{minipage}    
    \caption{Camera pose track and frames}
    \label{fig:track}
  \end{subfigure}
  \hfill
  \begin{subfigure}[b]{.33\textwidth}
    \centering\tiny
    \setlength{\figurewidth}{\textwidth}
    \setlength{\figureheight}{\textwidth}  
    \input{./fig/posekernel-cov.tex}
    \caption{Pose-kernel in batch mode}
    \label{fig:cov-batch}  
  \end{subfigure}
  \hfill
  \begin{subfigure}[b]{.33\textwidth}
    \centering\tiny
    \setlength{\figurewidth}{\textwidth}
    \setlength{\figureheight}{\textwidth}  
    \input{./fig/posekernel-cov-online.tex}
    \caption{Pose-kernel in chain mode (online)}
    \label{fig:cov-online}  
  \end{subfigure}
  \caption{(a)~A continuous camera trajectory on the left with associated camera frames. In (b)--(c), the {\it a~priori} pose-kernel covariance structures are shown as matrices (colormap: $0\,$\protect\includegraphics[width=8mm]{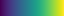}$\,\gamma^2$). The kernel encodes information about how much similarity (or correlation) we expect certain views to have in their latent space. See, \eg, the correlation between poses 1--4 and 9. In (b) this correlation is propagated over the entire track, while in (c) the long-range effects are suppressed. The small coordinate \textcolor{red}{x}\textcolor{green}{y}\textcolor{blue}{z}-axes illustrate camera orientations.}
  \label{fig:pose-kernel-cov}
\end{figure*}

In the encoder, there are five convolutional layers (a $7{\times}7$ filter for the first layer, a $5{\times}5$ filter for the second, and $3{\times}3$ filters for others).  After encoding, we get a latent-space representation $\vy$ of size $512 {\times} 8 {\times} 10$, which will be transformed by the GP model. Then decoder will take the transformed latent representation $\vz$ as the input to generate a $1{\times} H {\times} W$ prediction. There are four skip connections between the encoder and decoder and the inverse depth maps are predicted at four scales. All convolutional layers are followed by batch normalization and a ReLU function. The prediction layers using sigmoid function scaled by two to constrain the range of the predictions. To support arbitrary length of inputs,  when there are more than one neighbour frame, we compute the cost volume with each neighbour image separately and then average the cost volumes before passing them to the encoder--decoder network. 

During training, with a sequence of $N$ input frames, we predict $N$ depth maps by using the previous frame as the neighbour frame (except for the first frame that use the next frame as the neighbour frame), and use the mean of the L1 errors (at four scales) of all frames as the overall loss for training the model.

\begin{figure*}[!t]
  \centering
  \setlength{\figurewidth}{.140\textwidth}
  \setlength{\figureheight}{0.75\figurewidth}

  \newcommand{\figg}[1]{\includegraphics[width=.98\figurewidth]{./fig/comparison/#1}}

  \newcommand{\figrow}[2]{%
     \node [draw=white,thick,minimum width=\figurewidth,inner sep=0] at
       ({0*\figurewidth},#2) {\figg{#1-ref.jpg}};%
     \node [draw=white,thick,minimum width=\figurewidth,inner sep=0] at
       ({1*\figurewidth},#2) {\figg{#1-gt.jpg}};%
     \node [draw=white,thick,minimum width=\figurewidth,inner sep=0] at
       ({2*\figurewidth},#2) {\figg{#1-ours.jpg}};%
     \node [draw=white,thick,minimum width=\figurewidth,inner sep=0] at
       ({3*\figurewidth},#2) {\figg{#1-mvdepth.jpg}};%
     \node [draw=white,thick,minimum width=\figurewidth,inner sep=0] at
       ({4*\figurewidth},#2) {\figg{#1-deepmvs.jpg}};%
     \node [draw=white,thick,minimum width=\figurewidth,inner sep=0] at
       ({5*\figurewidth},#2) {\figg{#1-mvsnet.jpg}};%
     \node [draw=white,thick,minimum width=\figurewidth,inner sep=0] at
       ({6*\figurewidth},#2) {\figg{#1-colmap.jpg}};%
  }

  \begin{tikzpicture}

  \def\myarray{{"reference","ground-truth","ours (batch)","mvdepthnet","deepmvs","mvsnet","colmap"}}
  \foreach \i in {0,...,6}
     \node[text width=.9\figurewidth,align=center,text centered,text depth = 0cm] at ({\figurewidth*\i},{-.65*\figureheight}) {\scriptsize \sc \vphantom{$^\dagger$}\pgfmathparse{\myarray[\i]}\pgfmathresult};

  \figrow{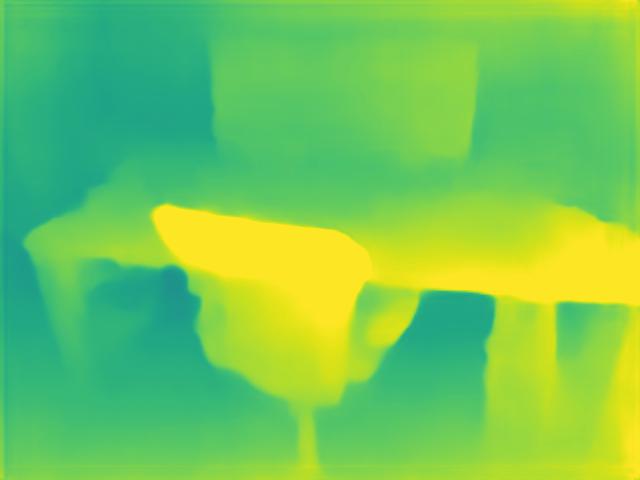}{0}  
  \figrow{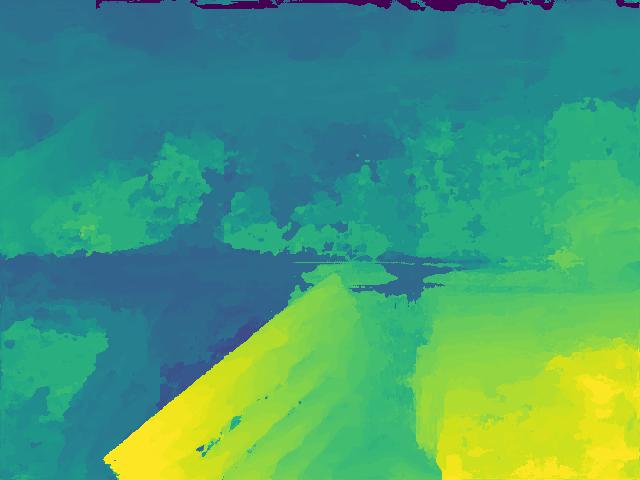}{1.0\figureheight}
  \figrow{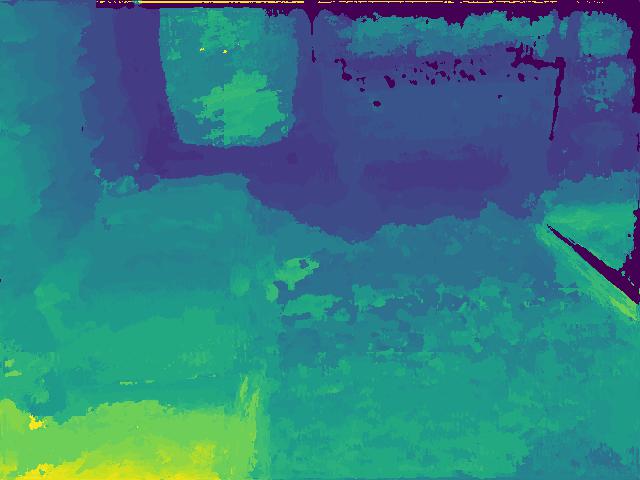}{2.0\figureheight}

  \figrow{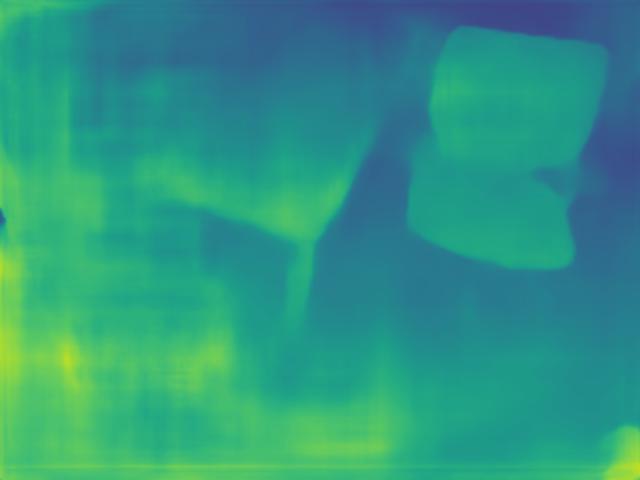}{3.0\figureheight}
        
  \end{tikzpicture}   
  \caption{Qualitative comparisons on the \textsc{Sun3d} and \textsc{7scenes}.}
  \label{fig:frames}
\end{figure*}
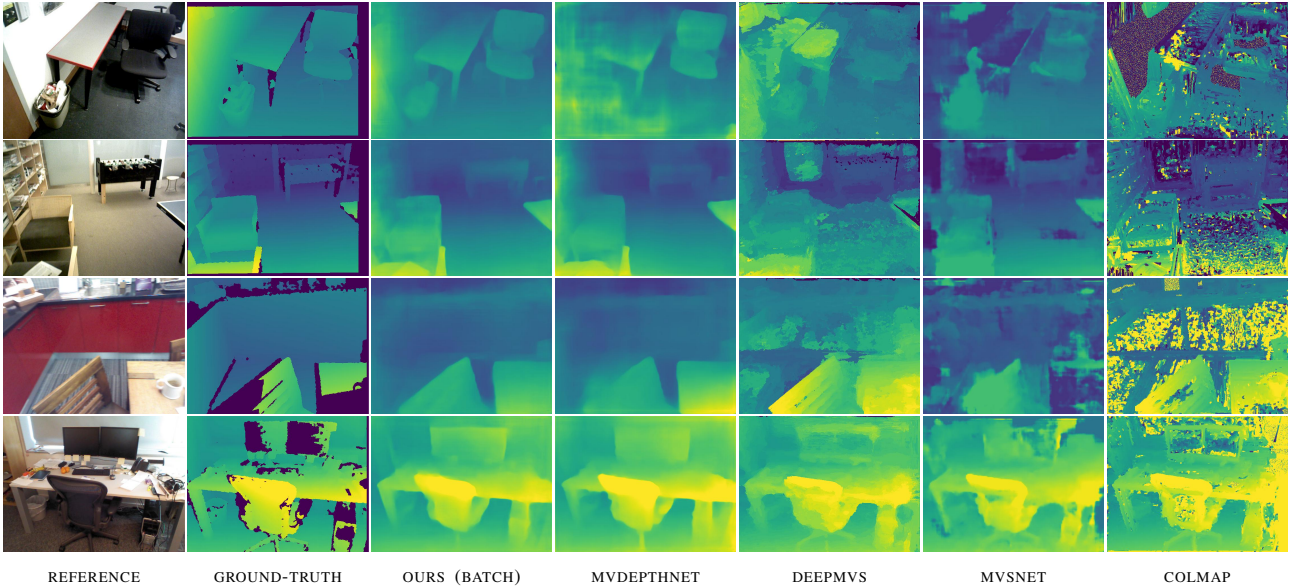

\subsection{Pose-kernel Gaussian process prior}
\label{sec:pose-kernel}
\noindent
We seek to define a probabilistic prior on the latent space that would account for {\it a~priori} knowledge that poses with close or overlapping field of view should produce more similar latent space encodings than poses far away from each other or poses with the camera pointing in opposite directions. This knowledge is to be encoded by a covariance function (kernel), and for this we need to define a distance measure or {\em metric} to define `closeness' in pose-space.

To measure the distance between camera poses, we build upon the work by Mazzotti~\etal \cite{mazzotti2016measure} which considers measures of rigid body poses. We extend this work to be suitable for computer vision applications. Specifically, we propose the following pose-distance measure between two camera poses $P_i$ and $P_j$:
\begin{equation}\label{eq:distance}
  D[P_i,P_j] = \sqrt{\|\vt_i-\vt_j\|^2 + \frac{2}{3}\,\mathrm{tr}(\MI - \MR_i^\T \MR_j)},
\end{equation}
where the poses are defined as $P = \{\vt, \MR\}$ residing in $\mathbb{R}^3 \times \mathrm{SO}(3)$, $\MI$ is an identity matrix, and `$\mathrm{tr}$' denotes the matrix trace operator.

We define a covariance (kernel) function for the latent space bottleneck layer in Fig.~\ref{fig:architecture}. We design the prior for the latent space processes such that they are stationary and both mean square continuous and once differentiable (see \cite{Rasmussen+Williams:2006}, Ch.~4) in pose-distance. This design choice is motivated by the fact that we expect the latent functions to model more structural than purely visual features, and that the we want the latent space to behave in a continuous and relatively smooth fashion. Choosing the covariance function structure from the so-called Mat\'ern class \cite{Rasmussen+Williams:2006} fulfils these requirements:
\begin{equation}\label{eq:matern}
  \kappa(P,P') = \gamma^2\,\bigg(1 + \frac{\sqrt{3}\, D[P,P']}{\ell} \bigg)\exp\!\bigg(-\frac{\sqrt{3}\, D[P,P']}{\ell}\bigg).
\end{equation}
This kernel encodes two arbitrary camera poses $P$ and $P'$ to `nearness' or similarity in latent space values subject to the distance (in the sense of Eq.~\ref{eq:distance}) of the camera poses. The tunable (learnable) hyperparameters $\gamma^2$ and $\ell$ define the characteristic magnitude and length-scale of the processes. Fig.~\ref{fig:pose-kernel-cov} shows an example camera pose track and associated covariance matrix evaluated from Eq.~\eqref{eq:matern} with unit hyperparameters.

In order to share the temporal information between frames in the sequence, we assign {\em independent} GP priors to all values in $\vz_i$, and consider the encoder outputs $\vy_i$ to be noise-corrupted versions of the `ideal' latent space encodings (see Fig.~\ref{fig:architecture}). This inference problem can be stated as the following GP regression model:
\begin{equation}
\begin{split}\label{eq:gp}
  z_j(t) &\sim \mathrm{GP}(0, \kappa(P[t],P[t'])), \\
  y_{j,i} &= z_j(t_i) + \varepsilon_{j,i}, \quad  \varepsilon_{j,i} \sim \mathrm{N}(0, \sigma^2),
\end{split}
\end{equation}
where $z_j(t)$, $j=1,2,\ldots,(512{\times}8{\times}10)$, are the values of the latent function $\vz$ at time $t$. The noise variance $\sigma^2$ is a parameter of the likelihood model, and thus the third and final free parameter to be learned.

\subsection{Latent-state batch estimation}
\label{sec:gp-batch}
\noindent
We first consider a batch solution for solving the inference problem in Eq.~\eqref{eq:gp} for an unordered set of image-pose pairs. Because the likelihood is Gaussian and all the GPs share the same poses at which the covariance function is evaluated, we may solve all the $512{\times}8{\times}10$ GP regression problems with one matrix inversion. This is due to the posterior covariance only being a function of the input poses, not values of learnt representations of images (\ie, $\vy$ does not appear in the posterior variance terms in Eq.~\ref{eq:gp-regression}). The posterior mean and covariance will be given by \cite{Rasmussen+Williams:2006}:
\begin{equation}
\begin{split}\label{eq:gp-regression}
  \!\!\!\!\mathbb{E}[\MZ \mid \{(P_i,\vy_i)\}_{i=1}^N] &{=} \MC \, (\MC+\sigma^2\,\MI)^{-1}\,\MY, \\
  \!\!\!\!\mathbb{V}[\MZ \mid \{(P_i,\vy_i)\}_{i=1}^N] &{=} \diag(\MC - \MC \, (\MC+\sigma^2\,\MI)^{-1}\,\MC),
\end{split}
\end{equation}
where $\MZ = (\vz_1~\vz_2~\ldots~\vz_N)^\top$ are stacked latent space encodings, $\MY = (\vy_1~\vy_2~\ldots~\vy_N)^\top$ are outputs from the encoder, and the covariance matrix $\MC_{i,j} = \kappa(P_i,P_j)$ (see Fig.~\ref{fig:cov-batch} for an example). The posterior mean $\mathbb{E}[\vz_i \mid \{(P_i,\vy_i)\}_{i=1}^N]$ is then passed through the decoder to output the predicted disparity map.

This batch scheme considers all the inter-connected poses in the sequence, making it powerful. The downside is that the matrix $\MC$ grows with the number of input frames/poses, $N$, and the inference requires inverting the matrix---which scales cubically in the matrix size. This scheme is thus applicable only to sequences with up to some hundreds of frames.

\begin{figure*}[!t]
  \centering
  \setlength{\figurewidth}{.140\textwidth}
  \setlength{\figureheight}{0.6667\figurewidth}

  \newcommand{\figg}[1]{\includegraphics[width=.98\figurewidth]{./fig/comparison/#1}}

  \newcommand{\figrow}[2]{%
     \node [draw=white,thick,minimum width=\figurewidth,inner sep=0] at
       ({0*\figurewidth},#2) {\figg{#1-ref.jpg}};%
     \node [draw=white,thick,minimum width=\figurewidth,inner sep=0] at
       ({1*\figurewidth},#2) {\figg{#1-gt}};%
     \node [draw=white,thick,minimum width=\figurewidth,inner sep=0] at
       ({2*\figurewidth},#2) {\figg{#1-ours}};%
     \node [draw=white,thick,minimum width=\figurewidth,inner sep=0] at
       ({3*\figurewidth},#2) {\figg{#1-mvdepth}};%
     \node [draw=white,thick,minimum width=\figurewidth,inner sep=0] at
       ({4*\figurewidth},#2) {\figg{#1-deepmvs}};%
     \node [draw=white,thick,minimum width=\figurewidth,inner sep=0] at
       ({5*\figurewidth},#2) {\figg{#1-mvsnet}};%
     \node [draw=white,thick,minimum width=\figurewidth,inner sep=0] at
       ({6*\figurewidth},#2) {\figg{#1-colmap}};%
  }

  \begin{tikzpicture}

  \def\myarray{{"reference","ground-truth","ours (batch)","mvdepthnet","deepmvs","mvsnet","colmap"}}
  \foreach \i in {0,...,6}
     \node[text width=.9\figurewidth,align=center,text centered,text depth = 0cm] at ({\figurewidth*\i},{-.65*\figureheight}) {\scriptsize \sc \vphantom{$^\dagger$}\pgfmathparse{\myarray[\i]}\pgfmathresult};

  \figrow{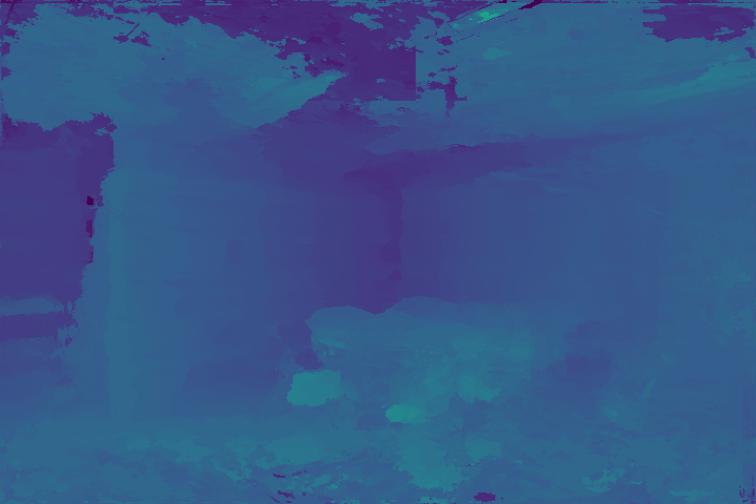}{0}  
  \figrow{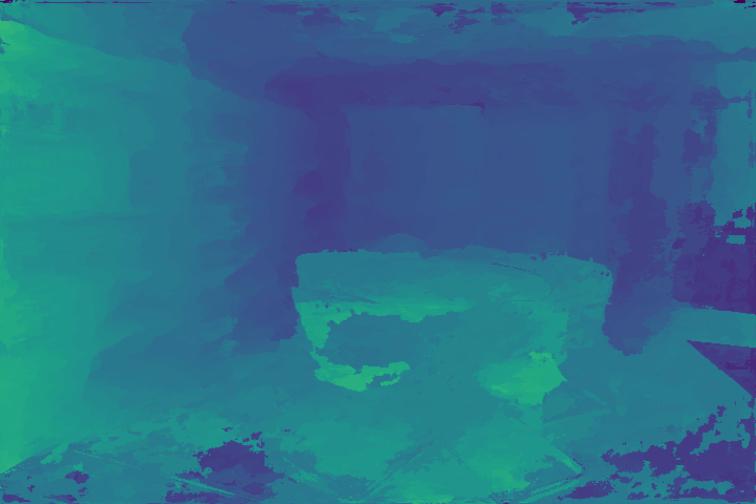}{1.03\figureheight}
  \figrow{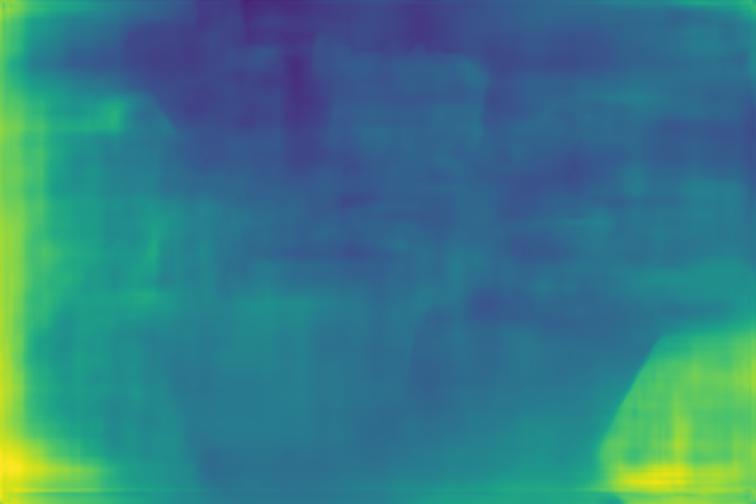}{2.06\figureheight}
  \figrow{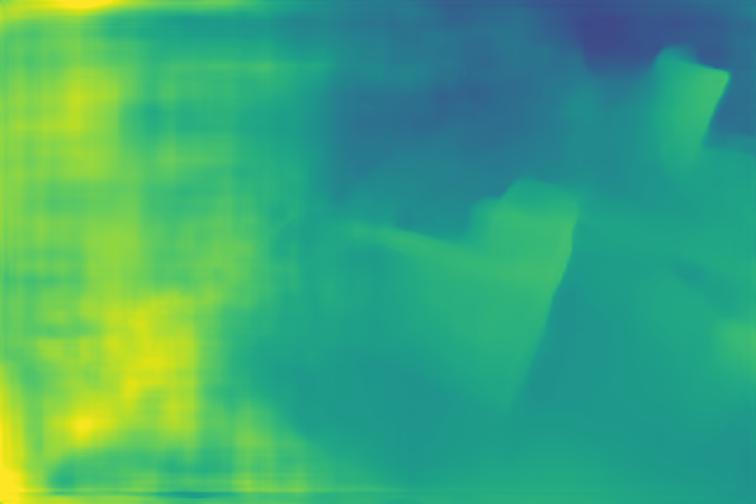}{3.09\figureheight}
        
  \end{tikzpicture}   
  \caption{Qualitative comparisons on the ETH3D dataset.}
  \label{fig:frames2}
\end{figure*}
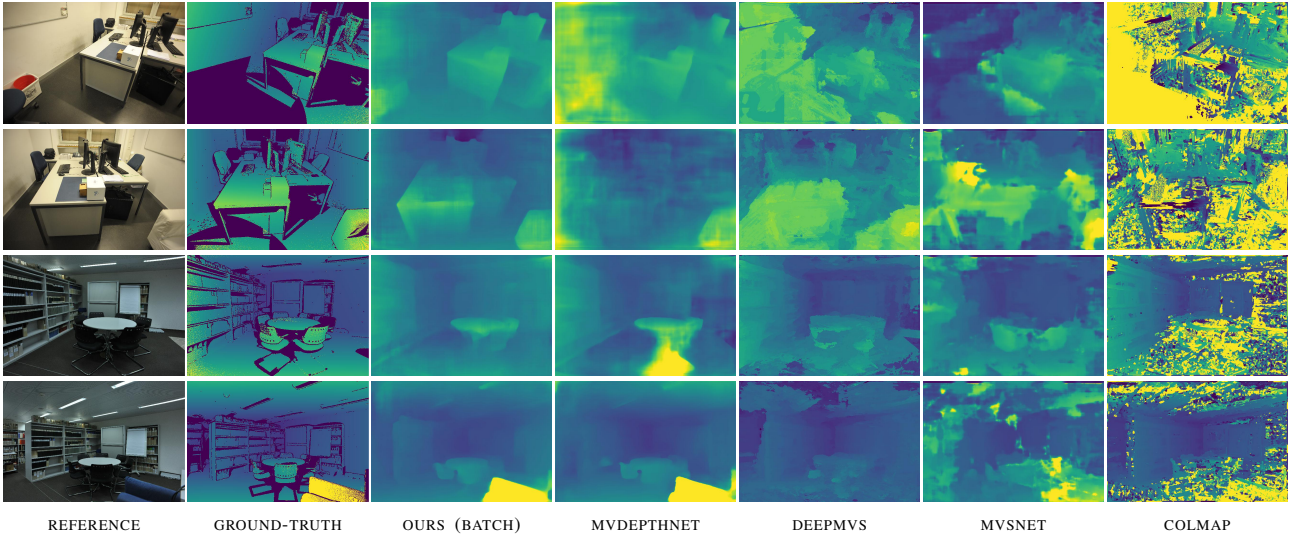

\subsection{Online estimation}
\label{sec:gp-online}
\noindent
In the case that the image-pose pairs have a natural ordering---as in a real-time application context---we may relax our model to a directed graph (\ie, Markov chain, see Fig.~\ref{fig:architecture} for the chain). In this case the GP inference problem can be solved in state-space form (see \cite{Sarkka+Solin+Hartikainen:2013,sarkka2019applied}) with a {\em constant} computational and memory complexity per pose/frame. The inference can be solved {\em exactly} without approximations by the following procedure \cite{Sarkka+Solin+Hartikainen:2013}.

For state-space GP inference, the covariance function (GP prior) is converted into a dynamical model. The initial (prior) state is chosen as the steady-state corresponding to the Mat\'ern covariance function (Eq.~\ref{eq:matern}): $\vz_0 \sim \mathrm{N}(\vmu_0, \MSigma_0)$, where $\vmu_0 = \vectb{0}$ and $\MSigma_0 = \diag(\gamma^2, 3\gamma^2/\ell^2)$. We jointly infer the posterior of all the independent GPs, such that the mean $\vmu_i$ is a matrix of size $2 {\times} (512{\cdot}8{\cdot}10)$, where the columns are the time-marginal means for the independent GPs and the two-dimensional state comes from the Mat\'ern model being once mean square differentiable. The covariance matrix is shared between all the independent GPs, $\MSigma_i \in \mathbb{R}^{2\times2}$. This makes the inference fast.

Following the derivation in \cite{Sarkka+Solin+Hartikainen:2013}, we define an evolution operator (which has the behaviour of the Mat\'ern)
\begin{equation} 
  \MPhi_i = \exp\!\bigg[\begin{pmatrix}0 & 1 \\ -3/\ell^2 & -2\sqrt{3}/\ell \end{pmatrix}  \Delta P_i \bigg],
\end{equation}
where the pose difference $\Delta P_i = D[P_i,P_{i-1}]$ is the pose-distance between consecutive poses. This gives us the predictive latent space values $\vz_i \mid \vy_{1:i-1}\sim \mathrm{N}(\bar\vmu_i,\bar\MSigma_i)$, where the mean and covariance are propagated by:
\begin{equation}
  \bar\vmu_i = \MPhi_i\,\vmu_{i-1}, \qquad
  \bar\MSigma_i = \MPhi_i \, \MSigma_{i-1} \, \MPhi_i^\T + \MQ_i,
\end{equation}
where $\MQ_i = \MSigma_0 - \MPhi_i \, \MSigma_{0} \, \MPhi_i^\T$. The posterior mean and covariance is then given by conditioning on the encoder output $\vy_i$ of the current step: 
\begin{equation}
\label{eq:kalman-filter}
  \!\!\!\!\!\vmu_{i} = \bar\vmu_{i} + \vk_i \, (\vy_i^\T - \vh^\T \bar\vmu_{i}), \quad
  \MSigma_{i} = \bar\MSigma_{i} - \vk_i \, \vh^\T \bar\MSigma_{i},
\end{equation}
where $\vk_i {=} \bar\MSigma_{i} \, \vh / (\vh^\T \bar\MSigma_{i} \, \vh + \sigma^2)$ and the observation model $\vh {=} (1~~0)^\T$. The posterior latent space encodings $\vz_i \mid \vy_{1:i} \sim \mathrm{N}(\vmu_{i},\MSigma_{i})$ conditioned on all image-pose pairs up till the current are then passed through the decoder to produce the disparity prediction. Due to overloaded notation (the state-space model tracks both the latent space values and their derivatives), it is actually $\vh^\T\!\vmu_{i}$ that is passed to the decoder.

\begin{figure*}[!t]
\begin{minipage}[c]{.67\linewidth}
 \captionof{table}{Comparison results between COLMAP, MVSNet, DeepMVS, MVDepthNet, and our method. We outperform other methods in most of the data sets and error metrics (smaller better).}
\label{tbl:results}
 {\noindent\footnotesize%
  \setlength{\tabcolsep}{7pt}
  \begin{tabular*}{\textwidth}{%
    >{\raggedright}p{0.1\textwidth} 
    P{0.1\textwidth} P{0.1\textwidth} P{0.1\textwidth} P{0.1\textwidth}  P{0.1\textwidth} P{0.1\textwidth}}
  \toprule
    & COLMAP &  MVSNet & DeepMVS & MVDepthNet  & Ours\,(online) & Ours\,(batch)  \\
\multicolumn{7}{l}{\bf SUN3D \hfill \rule[3pt]{.87\textwidth}{.1pt}} \\
~~L1-rel	 &   0.8169 &0.3971 & 0.4196& 0.1147&0.1064&\textbf{0.1010}\\
~~L1-inv	 &  0.5356  &0.1204 &0.1103 & 0.0610&0.0548&\textbf{0.0512}\\
~~sc-inv	& 0.8117 &0.3355 & 0.3288& 0.1320&0.1268&\textbf{0.1220}\\
~~L1         & 1.6324 &0.6538 & 0.9923& 0.2631&0.2512&\textbf{0.2386}\\
\multicolumn{7}{l}{\bf 7SCENES \hfill \rule[3pt]{.87\textwidth}{.1pt}} \\
~~L1-rel	  & 0.5923&0.2789 & 0.2198& 0.1972&0.1706&\textbf{0.1583}\\
~~L1-inv	& 0.4160& 0.1201 &  0.0946& 0.1064&0.0931&\textbf{0.0884}\\
~~sc-inv	& 0.4553 &0.2570& 0.2258& 0.1611&0.1490&\textbf{0.1458}\\
~~L1        & 1.0659 & 0.4971&  0.4183&0.3807&0.3187&\textbf{0.2947}\\
\multicolumn{7}{l}{\bf ETH3D \hfill \rule[3pt]{.87\textwidth}{.1pt}} \\
~~L1-rel	 &   0.5574& 0.4706& 0.4124& 0.2569&0.2354 &\textbf{0.2291}\\
~~L1-inv	 &   0.4307 & 0.1901& 0.3380& 0.1366&0.1227 &\textbf{0.1066}\\
~~sc-inv	&   0.5595& 0.4555&0.4661 & 0.2667&0.2561&\textbf{0.2517}\\
~~L1         & 0.6440 &0.9567 & 0.5684&0.5979&0.5417&\textbf{0.5374}\\
\bottomrule
  \end{tabular*}}
  
\end{minipage}
\hfill
\begin{minipage}[c]{.3\linewidth}
  \centering\footnotesize
  \begin{tikzpicture}

    \node at (0,0) {\includegraphics[width=0.65\columnwidth]{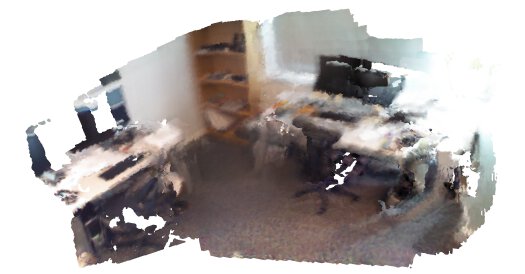}};
    \node at (0,-2) {\includegraphics[width=0.65\columnwidth]{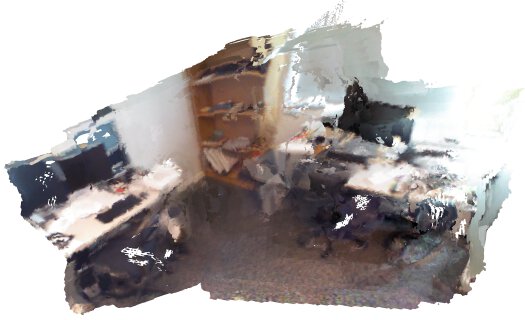}};
    \node at (0,-4) {\includegraphics[width=0.65\columnwidth]{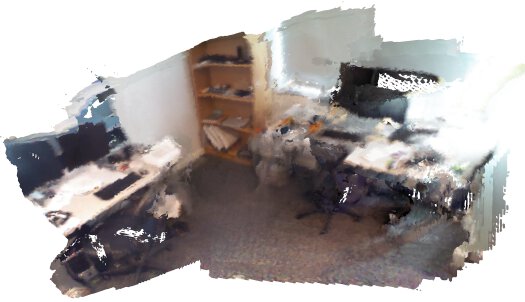}};

    \node[rotate=90] at (-2.3,0) {Ground-truth};
    \node[rotate=90] at (-2.3,-2) {Without GP};
    \node[rotate=90] at (-2.3,-4) {With GP};
    
  \end{tikzpicture}  
  \caption{3D reconstruction on \textsc{7scenes} by TSDF Fusion~\cite{zeng20163dmatch}. The results are fused from 25 depth maps. }
  \label{fig:tsdf}
\end{minipage}

\end{figure*}

\section{Experiments}
\label{sec:experiments}
\noindent
We train our model with the same data as in DeMoN~\cite{ummenhofer2017demon}. The training data set includes short sequences from real-world data sets \textsc{sun3d}~\cite{xiao2013sun3d}, RGBD~\cite{sturm2012benchmark}, MVS (includes \textsc{Citywall} and \textsc{Achteck-Turm}~\cite{fuhrmann2014mve}), and a synthesized data set \textsc{scenes11}~\cite{ummenhofer2017demon}. There are 92,558 training samples and each training sample consists of a three-view sequence with ground-truth depth maps and camera poses.  The resolution of the input images is $320{\times}256$. All data in our training set are also used in the training set of MVDepthNet, but the size of our training set is much smaller, so the improvement of the performance should not be explained by our training set. We load the MVDepthNet pretrained model as the starting point of training.  We jointly train the encoder, decoder, and the GP hyperparameters on a desktop worstation (NVIDIA GTX~1080~Ti, i7-7820X CPU, and 63~GB memory) using the Adam solver \cite{kingma2014adam} with $\beta_1 = 0.9$ and $\beta_2 = 0.999$, and a learning rate of  $10^{-4}$. The model was implemented in PyTorch and trained with 46k iterations with a mini-batch size of four. During training, we use the batch GP scheme (Sec.~\ref{sec:gp-batch}). After training, the GP hyperparameters were $\gamma^2 = 13.82$, $\ell = 1.098$, and $\sigma^2 = 1.443$.

\subsection{Evaluation}
\label{sec:eval}
\noindent
We evaluate our method on four sequences picked randomly from the indoor data set \textsc{7scenes}~\cite{shotton2013scene}  (\textit{office-01}, \textit{office-04},\textit{redkitchen-01}, \textit{redkitchen-02}). The \textsc{7scenes} data set can be regarded as an ideal evaluation data set as none of models are trained with \textsc{7scenes}, the results can reveal the generalization abilities of models. Moreover, sequences in \textsc{7scenes} generally contain different viewpoints in the same room, so there are many neighbour views that share similar scenes, which is applicable for studying the impact of our fusion scheme. Four sequences from \textsc{sun3d} (\textit{mit\_46\_6lounge}, \textit{mit\_dorm\_mcc\_eflr6}, \textit{mit\_32\_g725}, \textit{mit\_w85g})  and two sequences from \textsc{ETH3D} (\textit{kicker}, \textit{office}) are in evaluation of predicted depth maps. Altogether, there are 951 views in the evaluation set.

We use four common error metrics: {\em (i)} L1, {\em (ii)} L1-rel, {\em (iii)} L1-inv, and {\em (iv)} sc-inv.  The three L1 metrics are mean absolute difference, mean absolute relative difference, and mean absolute difference in inverse depth, respectively. They are given as
$\text{L1} = \frac{1}{n}\sum_i |d_i - \hat{d}_i|$, 
$\text{L1-rel} = \frac{1}{n}\sum_i \nicefrac{|d_i - \hat{d}_i|}{\hat{d}_i}$, and
$\text{L1-inv} = \frac{1}{n}\sum_i |d_i^{-1} - \hat{d}_i^{-1}|$,
where $d_i$ (meters) is the predicted depth value, $\hat{d}_i$ (meters) is the ground-truth value, $n$ is the number of pixels for which the depth is available. The scale-invariant metric is
  $\text{sc-inv} = (\frac{1}{n} \sum_i z_i^2 - \nicefrac{1}{n^2}(\sum_i z_i)^2)^{\nicefrac{1}{2}}$,
where $z_i = \log d_i - \log \hat{d}_i$. L1-rel normalizes the error, L1-inv puts more importance to close-range depth values, and sc-inv is a scale-invariant metric.

We compare our method with three state-of-the-art CNN-based MVS methods (MVSNet~\cite{yao2018mvsnet}, DeepMVS~\cite{huang2018deepmvs}, and MVDepthNet~\cite{wang2018mvdepthnet}), and one traditional MVS method (COLMAP, \cite{Schonberger+Zheng+Frahm+Pollefeys:2016}), because all these methods are available to image sequences.  For COLMAP, we use the ground-truth poses to produce dense models directly. For MVSNet, 192 depth labels based on ground-truth depth are used.  For COLMAP, MVSNet, and DeepMVS, to get good results, four neighbour images are assigned for each reference image, while MVDepthNet and our method only use one previous frame that has enough angle difference (${>}15^\circ$) or baseline translation (${>}0.1$~m) as the neighbour frame.

As shown in Table~\ref{tbl:results}, our methods, both the online and batch version, outperform other methods on all evaluation sets/metrics. Compared with the original MVDepthNet, the performance gets improved on all data sets after introducing the GP prior. These results underline, that sharing information across different poses always seems beneficial. As expected, the online estimation results are slightly worse than the batch estimation, because the online method only leverages the frames in the past. All models are trained with similar scenes, except for MVSNet which is trained with the DTU dataset that has much smaller scale of depth ranges; as our test sequences have larger ranges, the depth labels might become too sparse for the model, explaining its failed predictions. As also noted in the original publications, running COLMAP and DeepMVS is slow (orders of magnitude slower than the other methods). In comparison to MVDepthNet, as the GP inference only adds the cost of some comparably small matrix calculations which is small in comparison to the network evaluations, the improvements come at almost no cost.

Fig.~\ref{fig:frames} and \ref{fig:frames2} show qualitative comparison results. Patch-based methods like DeepMVS and COLMAP more easily to suffer from textureless regions and are more noisy. Compared to MVDepthNet, introducing the GP prior helps to obtain more stable depth maps with sharper edges. Fig.~\ref{fig:tsdf} reveals the temporal consistency of our method and proves that it is supplementary to traditional fusion methods.

\subsection{Ablation studies}
\label{sec:ablation}
\noindent
We have conducted several ablation studies for the design choices in our method.

\paragraph{Number of neighbour frames}
MVS methods typically use more than two input frames to reduce the noise in the cost volume. Our method can also use more than just a pair of inputs. Table~\ref{tbl:ablation1} shows the results on \textit{redkitchen-02}, where we compare our method to MVDepthNet and MVSNet. The use of more input frames improves all methods, but does not change the conclusions. Without the GP prior, even using five frames is inferior to our method with only two frames.

\begin{table*}[!tb]
\begin{minipage}{\textwidth}
  \caption{Ablation experiment: Performance comparison w.r.t.\ different number of input frames.}
 \label{tbl:ablation1}
\centering
  {\noindent\footnotesize
\begin{tabular*}{\textwidth}{>{\raggedright}p{0.12\textwidth}|P{0.07\textwidth} P{0.07\textwidth} P{0.07\textwidth}|P{0.07\textwidth} P{0.07\textwidth} P{0.07\textwidth}|P{0.07\textwidth} P{0.07\textwidth} P{0.07\textwidth}}
  \toprule
\multirow{2}{*}{} & 
\multicolumn{3}{c}{2 frames} & 
\multicolumn{3}{c}{3 frames} & 
\multicolumn{3}{c}{5 frames} \\
   Metric / Methods               & MVDepthNet & MVSNet & Ours            & MVDepthNet & MVSNet & Ours            & MVDepthNet & MVSNet & Ours            \\ \midrule
L1-rel            & 0.2009     & 0.3159 & \textbf{0.1615} & 0.1897     & 0.2665 & \textbf{0.1460} & 0.1734     & 0.2758 & \textbf{0.1429} \\ 
L1-inv            & 0.1161     & 0.1435 & \textbf{0.0979} & 0.1064     & 0.1244 & \textbf{0.0881} & 0.1028     & 0.1195 & \textbf{0.0850} \\ 
sc-inv            & 0.1866     & 0.3250 & \textbf{0.1729} & 0.1809     & 0.2902 & \textbf{0.1598} & 0.1766     & 0.2809 & \textbf{0.1587} \\ 
L1                & 0.4238     & 0.6036 & \textbf{0.3386} & 0.3922     & 0.5133 & \textbf{0.3066} & 0.3619     & 0.5116 & \textbf{0.2964}\\
\bottomrule
\end{tabular*}}
\end{minipage}\\[1em]
\begin{minipage}[b]{\columnwidth}
  \captionof{table}{Performance comparison w.r.t.\ thresholds of translation.}
 \label{tbl:ablation2}
\centering
  {\noindent\footnotesize
\begin{tabular*}{\columnwidth}{>{\raggedright}p{0.25\textwidth} P{0.13\textwidth}  P{0.13\textwidth}  P{0.13\textwidth}  P{0.13\textwidth}}
  \toprule
\multirow{2}{*}{} & \multicolumn{2}{c}{$t_{\text{min}}=0.1$ m}          & \multicolumn{2}{c}{$t_{\text{min}}=0.05$ m}          \\
   Metric / Methods               & w/o GP & Ours            & w/o GP& Ours          \\ \midrule
L1-rel               & 0.1474 & 0.1238     & 0.1535 & 0.1262    \\ 
L1-inv           & 0.0790&  0.0660    & 0.0828& 0.0669    \\ 
sc-inv             & 0.1436& 0.1315   & 0.1487 & 0.1334  \\ 
L1                & 0.3098 & 0.2609   & 0.3242& 0.2664     \\
\bottomrule
\end{tabular*}}
\end{minipage}
\hfill
\begin{minipage}[b]{\columnwidth}
  \caption{Performance comparison w.r.t.\ different kernels.}
 \label{tbl:ablation3}
  \centering
  {\noindent\footnotesize
\begin{tabular*}{\columnwidth}{>{\raggedright}p{0.25\textwidth} P{0.13\textwidth}  P{0.13\textwidth}  P{0.13\textwidth}  P{0.13\textwidth}}
\toprule \\
Metric / Methods  & L1-rel & L1-inv & sc-inv & L1 \\ \midrule
Mat\'ern    & 0.1298       & 0.0683        &0.1384        & 0.2769    \\
Exponential & 0.1376       &0.0703        &0.1417        & 0.2846  \\
TD kernel  & 0.1450     &  0.0745   & 0.1457   & 0.3041 \\
w/o GP &  0.1538 		&0.0824     &0.1507      & 0.3265 \\
\bottomrule
\end{tabular*}}
\end{minipage}

\end{table*}

\paragraph{Neighbour selection}
Strict view selection rules are required in many methods to obtain good predictions, because the cost volume breaks down if there is not enough baseline between views. We study robustness by decreasing the threshold of translation when selecting the neighbour frame in the \textsc{sun3d} and \textsc{7scenes} sequences. In Table~\ref{tbl:ablation2}, the error metrics increase more without using GP priors, which signals that the GP is beneficial in cases where the camera does not move much.

\paragraph{Choice of kernel function}
In addition to the Mat\'ern kernel, we experiment with the exponential kernel \cite{Rasmussen+Williams:2006}:
  $\kappa(P,P') = \gamma^2\exp(-D[P,P']/\ell)$.
The exponential kernel does not encode any smoothness (not differentiable), which makes it too flexible for the task as can be read from the error metrics in Table~\ref{tbl:ablation3}. If we ignore the pose information in the kernel, and only use the temporal difference (TD) instead, $D[i, j] = | i - j |$, the GP can be seen as a low-pass filter. We experimented with the TD in the Mat\'ern kernel, which gives better results than not using a GP, but performed worse than both the GPs that use the pose-distance.

\subsection{Online experiments with iOS}
\begin{figure}
  \centering
  \tikz\node[draw=mycolor0,rounded corners=2pt]{\includegraphics[width=.95\columnwidth]{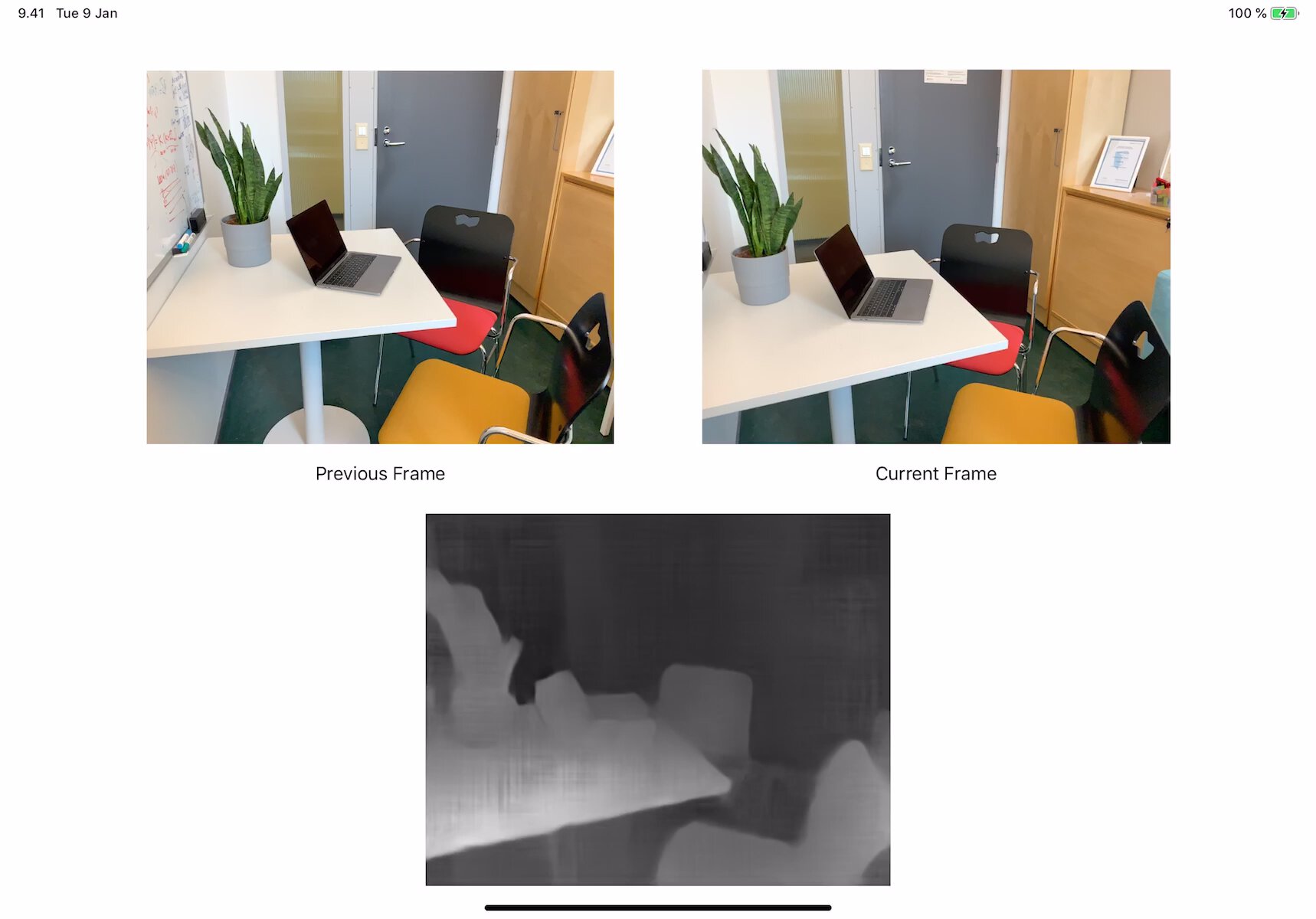}};
  \caption{Screenshot of our disparity estimation method running on an Apple iPad Pro (11-inch, late-2018 model). The previous and current frames are side-by-side on the top. The predicted disparity (corresponding to the current frame) is visualized on the bottom. The  pose information comes from Apple's ARKit API.}
  \label{fig:ios}
\end{figure}

\noindent
To demonstrate the practical value of our MVS scheme, we ported our implementation to an iOS app. The online GP scheme and cost volume construction were implemented in C++ with wrappers in Objective-C, while the app itself was implemented in Swift. More specifically, the homography warping for cost volume construction leverages OpenCV, and the real-time GP is implemented using the Eigen matrix library. The trained PyTorch model was converted to a CoreML model through ONNX. The camera poses are captured by Apple ARKit. Fig.~\ref{fig:ios} shows a screenshot of the app in action, where we have set the refresh rate to ${\sim}1$~Hz. Note that the model was not trained with any iOS data, nor any data from the environment the app was tested in.

\section{Discussion and conclusion}
\label{sec:discussion}
\noindent
In this paper, we proposed a novel idea for MVS that enables the model to leverage multi-view information but keep the frame structure simple and time-efficient at the same time. Our pose-kernel measures the `closeness' between frames and encodes this prior information using a Gaussian process in the latent space. In the experiments, we showed that this method clearly advances the state-of-the-art.
Our proposed model consistently improves the accuracy of estimated depth maps when appended to the baseline disparity network of \cite{wang2018mvdepthnet} and this holds independently of the number of views used for computing individual cost volumes. In addition, as the proposed model fuses information in the latent space, it is complementary with depth map fusion techniques, such as \cite{zeng20163dmatch}, which can fuse information from multiple depth maps estimated using our approach. In fact, besides improving individual depth map predictions, our latent-space GP prior leads to improved result also when combined with a subsequent depth map fusion stage.

One possible limitation of our method is that wrong predictions might also be propagated forward because of the fusion in the latent space. We do not employ any outlier rejection rules like traditional depth fusion methods. The same applies to occlusion. Even though we recognize this concern, we did not notice any problems in robustness while experimenting with our online app implementation. Still, introducing confidence measures to penalize wrong predictions could improve the method in the future.

Codes and material available on the project page:\\ \url{https://aaltoml.github.io/GP-MVS}.

\paragraph{Acknowledgements}
This research was supported by the Academy of Finland grants 308640, 324345, 277685, and 295081.  We acknowledge the computational resources provided by the Aalto Science-IT project.

{\small
\bibliographystyle{ieee_fullname}
\bibliography{bibliography}
}

\clearpage
\appendix

\twocolumn[
  \vspace*{1em}
  \begin{center}
  \Large\bf Supplementary material for\\
  Multi-View Stereo by Temporal Nonparametric Fusion
  \end{center}
  \vspace*{2em}
]

\section{Encoder--decoder architecture}
\noindent
We have included details on the network architecture that was used for the encoder and decoder models. Table~\ref{tbl:architecture} lists the netowrk components. `Ch.~I/O' refers to the channel number of the input/output. All `*\_up' means the upsampled features, and the upsample layers use bilinear interpolation. The plus sign `+' refers to the concatenation operation. The encoder consists of layers from `conv1' to 'conv5\_1' , and the output of the `conv5\_1' layer is $\vz$ in Fig.~\ref{fig:architecture}, which will be transformed by the GP. The layers from `upconv4' to `disp0' are part of the decoder, and the `disp0' generates the final inverse depth prediction. All layers are followed by batch normalization and ReLU, except the `disp*' layers. 

\begin{figure*}[!b]
  \resizebox{\textwidth}{!}{\def\ConvColor{rgb:yellow,5;red,2.5;white,5}
\def\ConvReluColor{rgb:yellow,5;red,5;white,5}
\def\PoolColor{rgb:red,1;black,0.3}
\def\UnpoolColor{rgb:blue,2;green,1;black,0.3}
\def\FcColor{rgb:blue,5;red,2.5;white,5}
\def\FcReluColor{rgb:blue,5;red,5;white,4}
\def\SoftmaxColor{rgb:magenta,5;black,7}   

\newcommand{\copymidarrow}{\tikz \draw[-Stealth,line width=0.8mm,draw={rgb:blue,4;red,1;green,1;black,3}] (-0.3,0) -- ++(0.3,0);}

%\begin{document}
\begin{tikzpicture}
\Large
\tikzstyle{connection}=[ultra thick,every node/.style={sloped,allow upside down},draw=\edgecolor,opacity=0.7]
\tikzstyle{copyconnection}=[ultra thick,every node/.style={sloped,allow upside down},draw={rgb:blue,4;red,1;green,1;black,3},opacity=0.7]

\tikzset{->-/.style={decoration={
  markings,
  mark=at position .5 with {\arrow{>}}},postaction={decorate}}}

\pic[shift={(0,0,0)}] at (0,0,0) 
    {Box={
        name=cv,
        caption= Input,
        xlabel={{67, }},
        zlabel=320,
        fill={rgb:red,4;green, 4;blue,5},
        height=40,
        width=3,
        depth=40
        }
    };

\pic[shift={ (2,0,0) }] at (cv-east) 
    {RightBandedBox={
        name=ccr_b1,
        caption=Conv1,
        xlabel={{ 128, 128 }},
        zlabel=160,
        fill=\ConvColor,
        bandfill=\ConvReluColor,
        height=40,
        width={ 5 , 5 },
        depth=40
        }
    };

\pic[shift={ (2,0,0) }] at (ccr_b1-east) 
    {RightBandedBox={
        name=b2,
        caption=Conv2,
        xlabel={{ 256, 256 }},
        zlabel=80,
        fill=\ConvColor,
        bandfill=\ConvReluColor,
        height=32,
        width={ 7.5 , 7.5 },
        depth=32
        }
    };

\pic[shift={ (2,0,0) }] at (b2-east) 
    {RightBandedBox={
        name=b3,
        caption=Conv3,
        xlabel={{ 512, 512 }},
        zlabel=40,
        fill=\ConvColor,
        bandfill=\ConvReluColor,
        height=25,
        width={ 10 , 10 },
        depth=25
        }
    };

\pic[shift={ (2,0,0) }] at (b3-east) 
    {RightBandedBox={
        name=b4,
        caption=Conv4,
        xlabel={{ 512, 512 }},
        zlabel=20,
        fill=\ConvColor,
        bandfill=\ConvReluColor,
        height=16,
        width={ 10 , 10 },
        depth=16
        }
    };

\pic[shift={ (2,0,0) }] at (b4-east) 
    {RightBandedBox={
        name=ccr_b5,
        caption=Conv5,
        xlabel={{ 512, 512 }},
        zlabel=10,
        fill=\ConvColor,
        bandfill=\ConvReluColor,
        height=8,
        width={ 10 , 10 },
        depth=8
        }
    };

\pic[shift={(2,0,0)}] at (ccr_b5-east) 
    {RightBandedBox={
        name=b6,
        caption=upconv4,
        xlabel={{512, }},
        zlabel=20,
        fill={rgb:red,1;green, 2;blue,5},
	bandfill={rgb:white,1;black,2},
        height=8,
        width=10,
        depth=8
        }
    };

\pic[shift={(2,0,0)}] at (b6-east) 
    {RightBandedBox={
        name=b7,
        caption=upconv3,
        xlabel={{512, }},
        zlabel=40,
        fill={rgb:red,1;green, 2;blue,5},
	bandfill={rgb:white,1;black,2},
        height=16,
        width=10,
        depth=16
        }
    };

\pic[shift={(5.7,0,15)}] at (b7-south) 
    {Box={
        name=disp3,
        caption=disp3,
        xlabel={{1, }},
        zlabel=40,
        fill=\SoftmaxColor,
        height=16,
        width=1,
        depth=16
        }
    };

\pic[shift={(3,0,0)}] at (b7-east) 
    {RightBandedBox={
        name=b8,
        caption=upconv2,
        xlabel={{256, }},
        zlabel=80,
        fill={rgb:red,1;green, 2;blue,5},
	bandfill={rgb:white,1;black,2},
        height=25,
        width=7.5,
        depth=25
        }
    };

\pic[shift={(5.7,0,15)}] at (b8-south) 
    {Box={
        name=disp2,
        caption=disp2,
        xlabel={{1, }},
        zlabel=80,
        fill=\SoftmaxColor,
        height=25,
        width=1,
        depth=25
        }
    };

\pic[shift={(3,0,0)}] at (b8-east) 
    {RightBandedBox={
        name=b9,
        caption=upconv1,
        xlabel={{128, }},
        zlabel=160,
        fill={rgb:red,1;green, 2;blue,5},
	bandfill={rgb:white,1;black,2},
        height=32,
        width=5,
        depth=32
        }
    };

\pic[shift={(5.7,0,15)}] at (b9-south) 
    {Box={
        name=disp1,
        caption=disp1,
        xlabel={{1, }},
        zlabel=160,
        fill=\SoftmaxColor,
        height=32,
        width=1,
        depth=32
        }
    };

\pic[shift={(3,0,0)}] at (b9-east) 
    {RightBandedBox={
        name=b10,
        caption=upconv0,
        xlabel={{64, }},
        zlabel=320,
        fill={rgb:red,1;green, 2;blue,5},
	bandfill={rgb:white,1;black,2},
        height=40,
        width=3,
        depth=40
        }
    };

\pic[shift={(5.7,0,15)}] at (b10-south) 
    {Box={
        name=disp0,
        caption=disp0,
        xlabel={{1, }},
        zlabel=320,
        fill=\SoftmaxColor,
        height=40,
        width=1,
        depth=40
        }
    };

\draw [connection]  (cv-east)    -- node {\midarrow} (ccr_b1-west);
\draw [connection]  (ccr_b1-east)    -- node {\midarrow} (b2-west);
\draw [connection]  (b2-east)    -- node {\midarrow} (b3-west);
\draw [connection]  (b3-east)    -- node {\midarrow} (b4-west);
\draw [connection]  (b4-east)    -- node {\midarrow} (ccr_b5-west);

\draw [connection]  (ccr_b5-east)    -- node {\midarrow} (b6-west);
\draw [connection]  (b6-east)    -- node {\midarrow} (b7-west);
\draw [connection]  (b7-east)    -- node {\midarrow} (b8-west);
\draw [connection]  (b8-east)    -- node {\midarrow} (b9-west);
\draw [connection]  (b9-east)    -- node {\midarrow} (b10-west);

% \draw [copyconnection]  (b2-northeast) -- ++(0,2cm) --node {\copymidarrow} ++(25cm,0) -| (b9-north);

\path (b4-east) -- (ccr_b5-west) coordinate[pos=0.25] (between4_5) ;
\path (b6-south)  -- (b6-north)  coordinate[pos=1.25] (b6-top) ;
\draw [copyconnection]  (between4_5)  
|- node {}(b6-top)
-- node {\copymidarrow} (b6-north);

\path (b3-east) -- (b4-west) coordinate[pos=0.25] (between3_4) ;
\path (b7-south)  -- (b7-north)  coordinate[pos=1.25] (b7-top) ;
\draw [copyconnection]  (between3_4)  
|- node {}(b7-top)
-- node {\copymidarrow} (b7-north);

\path (b2-east) -- (b3-west) coordinate[pos=0.25] (between2_3) ;
\path (b8-south)  -- (b8-north)  coordinate[pos=1.25] (b8-top) ;
\draw [copyconnection]  (between2_3)  
|- node {}(b8-top)
-- node {\copymidarrow} (b8-north);

\path (ccr_b1-east) -- (b2-west) coordinate[pos=0.25] (between1_2) ;
\path (b9-south)  -- (b9-north)  coordinate[pos=1.25] (b9-top) ;
\draw [copyconnection]  (between1_2)  
|- node {}(b9-top)
-- node {\copymidarrow} (b9-north);

\draw [connection]  (b7-south)    -- node {\midarrow} (disp3-north);
\draw [connection]  (b8-south)    -- node {\midarrow} (disp2-north);
\draw [connection]  (b9-south)    -- node {\midarrow} (disp1-north);
\draw [connection]  (b10-south)    -- node {\midarrow} (disp0-north);

\draw [connection]  (disp3-east) -- ++(1,0) --node {\copymidarrow} ++(0, 5) -| (b8-southwest);
\draw [connection]  (disp2-east) -- ++(1,0) --node {\copymidarrow} ++(0, 5) -| (b9-southwest);
\draw [connection]  (disp1-east) -- ++(1,0) --node {\copymidarrow} ++(0, 5) -| (b10-southwest);

\end{tikzpicture}
%\end{document}
}
  \caption{The architecture of the encoder--decoder in our method. The orange blocks are parts of the encoder, and the blue blocks are the decoder. The purple blocks indicate four ‘disp*' layers. Except ‘disp*' layers, each block are followed by a darker block which refer to the batch normalization and ReLU layers.}
  \label{fig:network-blocks}
\end{figure*}
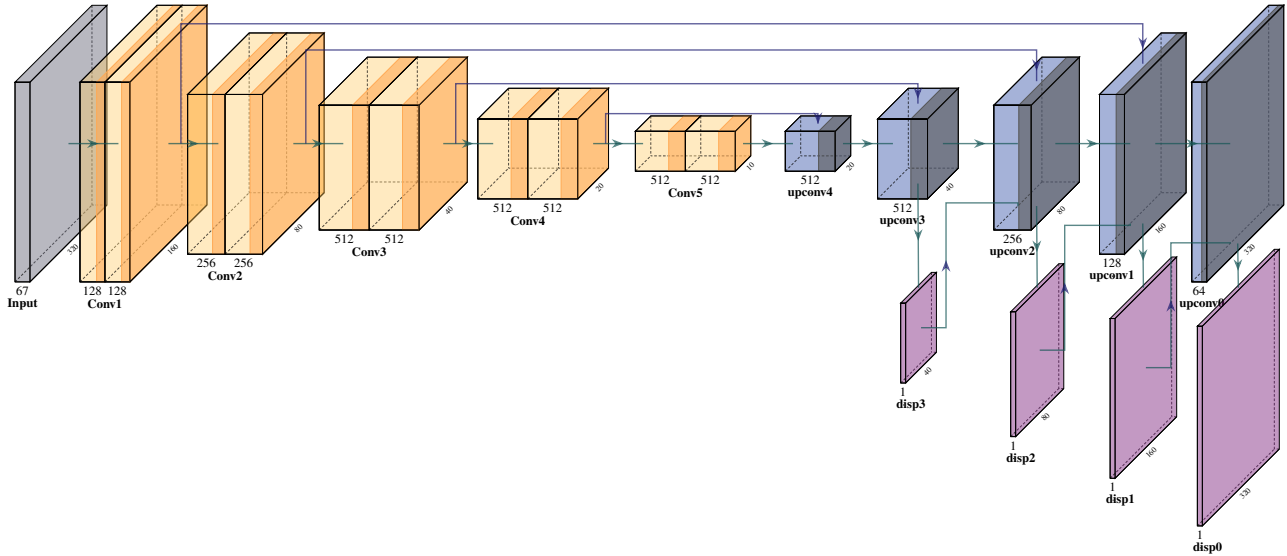

Additionally, Fig.~\ref{fig:network-blocks} visualizes the encoder--decoder arcitecture as a block diagram. The orange blocks are parts of the encoder, and the blue blocks form the decoder. The purple blocks indicate four ‘disp*' layers. Except ‘disp*' layers, each block is followed by a darker block which indicate batch normalization and ReLU layers. There are four skip connections in the figure, which corresponds to feeding the outputs of `conv*\_1' layers into `iconv*' layers.

\begin{table}[!h]
 \caption{Details of the encoder--decoder network structures.}
 \label{tbl:architecture}
\footnotesize
\begin{tabular*}{\columnwidth}{l c c c l}
  \toprule
\textbf{Name} & \textbf{Kernel}         & \textbf{s} & \textbf{Ch.\ I/O} & \textbf{Input} \\ \midrule
conv1         & $7{\times}7$ & 1               & 67/128          & reference image + cost volume \\
conv1\_1      & $7{\times}7$  & 2               & 128/128         & conv1                         \\
conv2         & $5{\times}5$  & 1               & 128/256         & conv1\_1                      \\
conv2\_1      & $5{\times}5$ & 2               & 256/256         & conv2                         \\
conv3         & $3{\times}3$  & 1               & 256/512         & conv2\_1                      \\
conv3\_1      & $3{\times}3$ & 2               & 512/512         & conv3                         \\
conv4         & $3{\times}3$ & 1               & 512/512         & conv3\_1                      \\
conv4\_1      & $3{\times}3$& 2               & 512/512         & conv4                         \\
conv5         & $3{\times}3$ & 1               & 512/512         & conv4\_1                      \\
conv5\_1      & $3{\times}3$& 2               & 512/512         & conv5                         \\
\midrule
upconv4       & $3{\times}3$ & 1               & 512/512         & conv5\_1(after GP)\_up                 \\
iconv4        & $3{\times}3$& 1               & 1024/512        & conv4\_1+upconv4              \\
upconv3       & $3{\times}3$& 1               & 512/512         & iconv4\_up                        \\
iconv3        & $3{\times}3$ & 1               & 1024/512        & conv3\_1+upconv3              \\
disp3         & $3{\times}3$& 1               & 512/1           & iconv3                        \\
upconv2       & $3{\times}3$ & 1               & 512/256         & iconv3\_up                     \\
iconv2        & $3{\times}3$ & 1               & 513/256         & conv2\_1+upconv2+disp3\_up        \\
disp2         & $3{\times}3$ & 1               & 256/1           & iconv2                        \\
upconv1       & $3{\times}3$ & 1               & 256/128         & iconv2\_up                        \\
iconv1        & $3{\times}3$& 1               & 257/128         & conv1\_1+upconv1+disp2\_up        \\
disp1         & $3{\times}3$ & 1               & 128/1           & iconv1                        \\
upconv0       & $3{\times}3$ & 1               & 128/64          & iconv1\_up                        \\
iconv0        & $3{\times}3$ & 1               & 65/64           & upconv0+disp1\_up                 \\
disp0         & $3{\times}3$ & 1               & 64/1            & iconv0 \\
\bottomrule                      
\end{tabular*}
\end{table}

\section{Additional examples}
\begin{figure*}[!t]
  \centering
  \setlength{\figurewidth}{.140\textwidth}
  \setlength{\figureheight}{0.75\figurewidth}

  \newcommand{\figg}[1]{\includegraphics[width=.98\figurewidth]{./fig/comparison/#1}}

  \newcommand{\figrow}[2]{%
     \node [draw=white,thick,minimum width=\figurewidth,inner sep=0] at
       ({0*\figurewidth},#2) {\figg{#1-ref.jpg}};%
     \node [draw=white,thick,minimum width=\figurewidth,inner sep=0] at
       ({1*\figurewidth},#2) {\figg{#1-gt.jpg}};%
     \node [draw=white,thick,minimum width=\figurewidth,inner sep=0] at
       ({2*\figurewidth},#2) {\figg{#1-ours.jpg}};%
     \node [draw=white,thick,minimum width=\figurewidth,inner sep=0] at
       ({3*\figurewidth},#2) {\figg{#1-mvdepth.jpg}};%
     \node [draw=white,thick,minimum width=\figurewidth,inner sep=0] at
       ({4*\figurewidth},#2) {\figg{#1-deepmvs.jpg}};%
     \node [draw=white,thick,minimum width=\figurewidth,inner sep=0] at
       ({5*\figurewidth},#2) {\figg{#1-mvsnet.jpg}};%
     \node [draw=white,thick,minimum width=\figurewidth,inner sep=0] at
       ({6*\figurewidth},#2) {\figg{#1-colmap.jpg}};%
  }

  \begin{tikzpicture}

  \def\myarray{{"reference","ground-truth","ours (batch)","mvdepthnet","deepmvs","mvsnet","colmap"}}
  \foreach \i in {0,...,6}
     \node[text width=.9\figurewidth,align=center,text centered,text depth = 0cm] at ({\figurewidth*\i},{-.65*\figureheight}) {\scriptsize \sc \vphantom{$^\dagger$}\pgfmathparse{\myarray[\i]}\pgfmathresult};

  \figrow{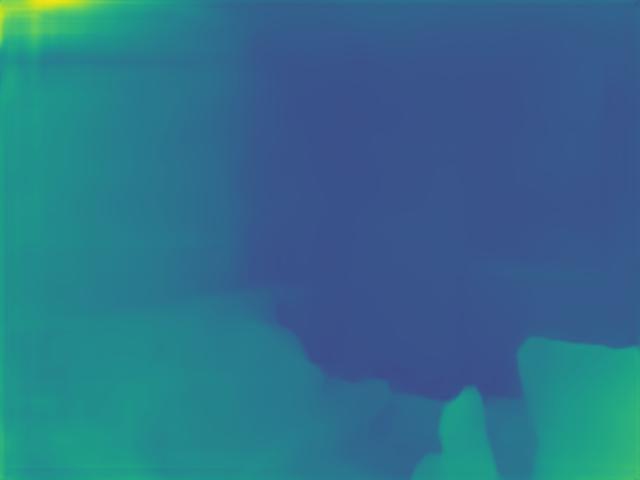}{0}  
  \figrow{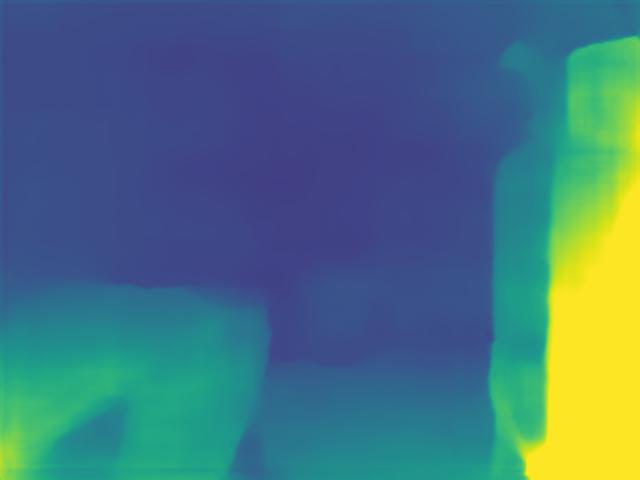}{1.02\figureheight}
  \figrow{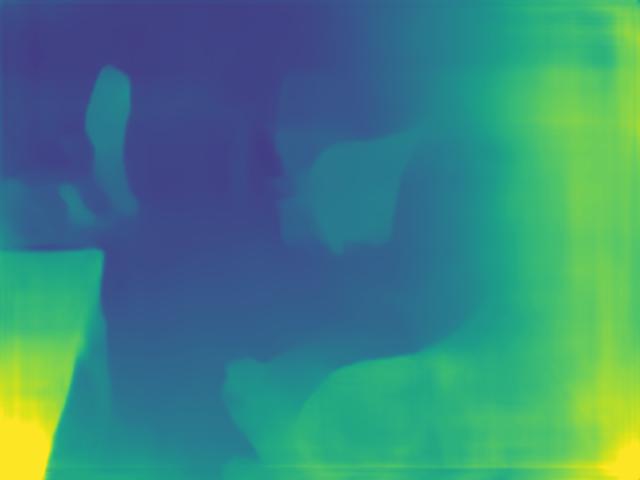}{2.04\figureheight}
  \figrow{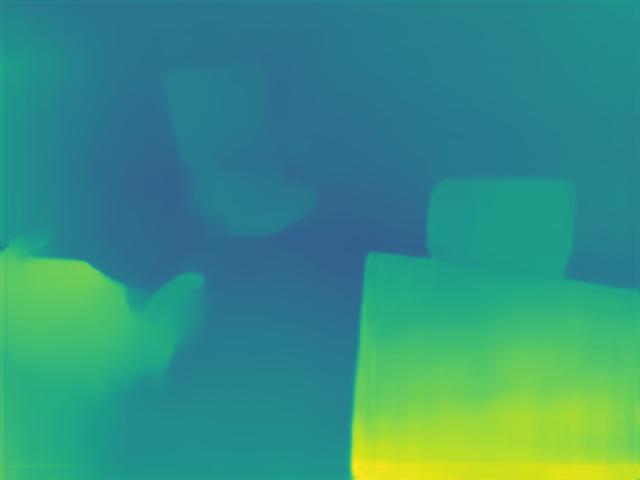}{3.06\figureheight}
  \figrow{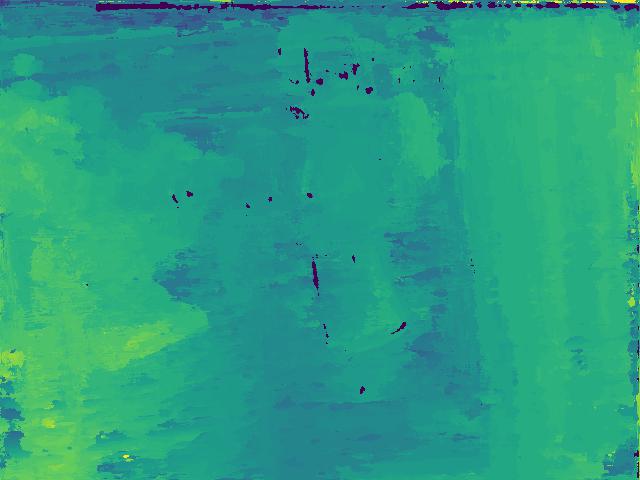}{4.08\figureheight}
  \figrow{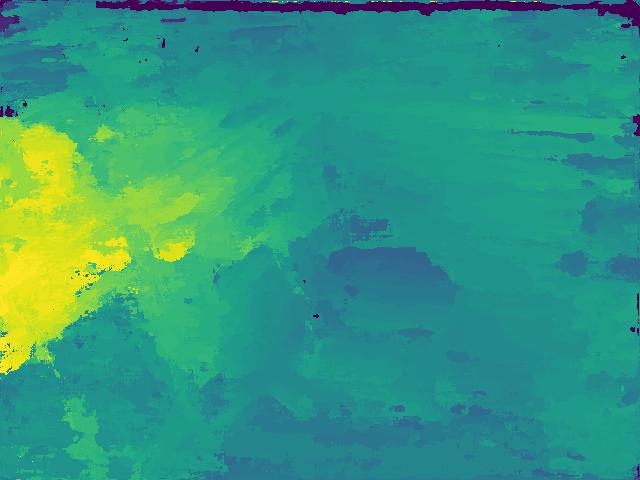}{5.10\figureheight}

  \end{tikzpicture}   
  \caption{Qualitative results on \textsc{SUN3D}.}
  \label{fig:addexamples}
\end{figure*}

\begin{figure*}[!t]
  \centering
  \setlength{\figurewidth}{.280\textwidth}
  \setlength{\figureheight}{0.75\figurewidth}

  \newcommand{\figg}[1]{\includegraphics[width=.98\figurewidth]{./fig/mesh/#1}}

  \newcommand{\figrow}[2]{%
     \node [draw=white,thick,minimum width=\figurewidth,inner sep=0] at
       ({0*\figurewidth}, #2) {\figg{#1-gt}};%
     \node [draw=white,thick,minimum width=\figurewidth,inner sep=0] at
       ( {1*\figurewidth}, #2) {\figg{#1-mvdepth}};%
     \node [draw=white,thick,minimum width=\figurewidth,inner sep=0] at
       ( {2*\figurewidth}, #2) {\figg{#1-ours}};%

  }

  \begin{tikzpicture}

  \def\myarray{{"ground-truth",
"w/o GP", "ours"}}
  \foreach \i in {0,...,2}
     \node[text width=.9\figurewidth,align=center,text centered,text depth = 0cm] at ({\figurewidth*\i},{-.65*\figureheight}) {\scriptsize \sc \vphantom{$^\dagger$}\pgfmathparse{\myarray[\i]}\pgfmathresult};

  \figrow{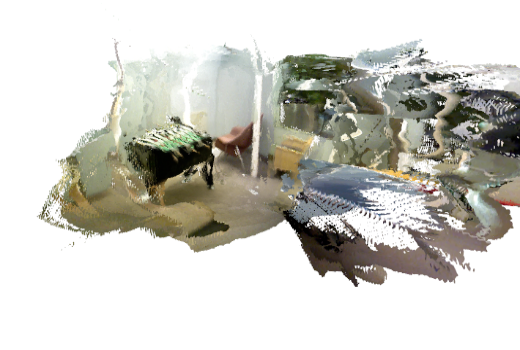}{0}  
  \figrow{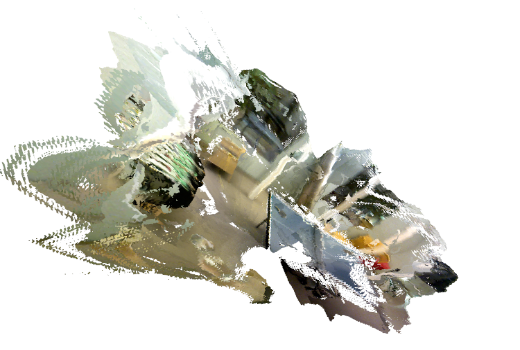}{1.03\figureheight}
  \figrow{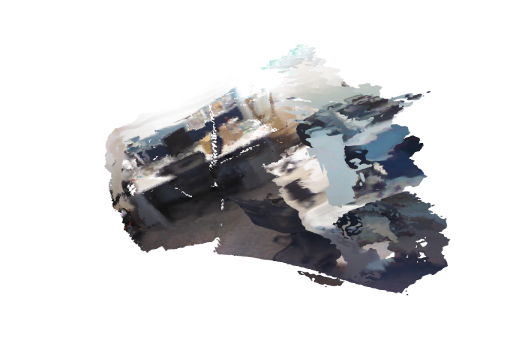}{2.06\figureheight}
  \figrow{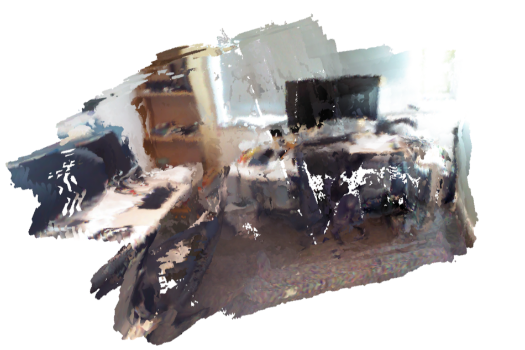}{3.09\figureheight}
  \figrow{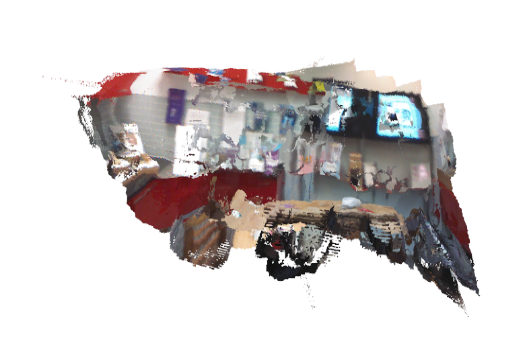}{4.12\figureheight}    
  \figrow{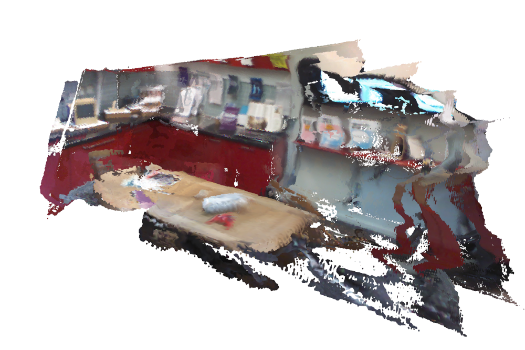}{5.15\figureheight}

  \end{tikzpicture}   
  \caption{3D reconstruction examples. All results are fused from 25 depth maps.}
  \label{fig:ply}
\end{figure*}

\noindent
In addition to those in the main paper, we show additional qualitative comparisons of our method and other methods in Fig.~\ref{fig:addexamples}. In these example frames, our method predicts noiseless dense depth maps with more details compared with other methods. For example, with our method, the shape of arms of chairs in row~3 and row~4 is more clearer. Moreover, our method provide more accurate prediction in both near and far parts of the scene. For instance, the bag in row~6 and the chair in row~5 show the better performance in close by areas, and the table in row~4 and the fridge in row~6 show the better performance in slightly farther areas. Fig. ~\ref{fig:ply} shows more 3D reconstruction results by applying TSDF fusion on 25 predicted depth maps, which prove that our method have better performance on temporal consistency. 

\begin{figure*}[!t]
  \centering
  \setlength{\figurewidth}{.180\textwidth}
  \setlength{\figureheight}{0.75\figurewidth}

  \newcommand{\figg}[1]{\includegraphics[width=.98\figurewidth]{./fig/fail/#1}}

  \newcommand{\figcol}[2]{%
     \node [draw=white,thick,minimum width=\figurewidth,inner sep=0] at
       (#2,{3*\figureheight}) {\figg{#1-ref.jpg}};%
     \node [draw=white,thick,minimum width=\figurewidth,inner sep=0] at
       (#2, {2*\figureheight}) {\figg{#1-gt}};%
     \node [draw=white,thick,minimum width=\figurewidth,inner sep=0] at
       (#2, {1*\figureheight}) {\figg{#1-ours}};%

  }

  \begin{tikzpicture}

  \def\myarray{{"ours","ground-truth","reference"}}
  \foreach \i in {1, 2, 3}
     \node[text width=.9\figurewidth,align=center,text centered,text depth = 0cm, rotate=90] at ({\figurewidth*-.65},{\i*\figureheight}) {\scriptsize \sc \vphantom{$^\dagger$}\pgfmathparse{\myarray[\i-1]}\pgfmathresult};

  \figcol{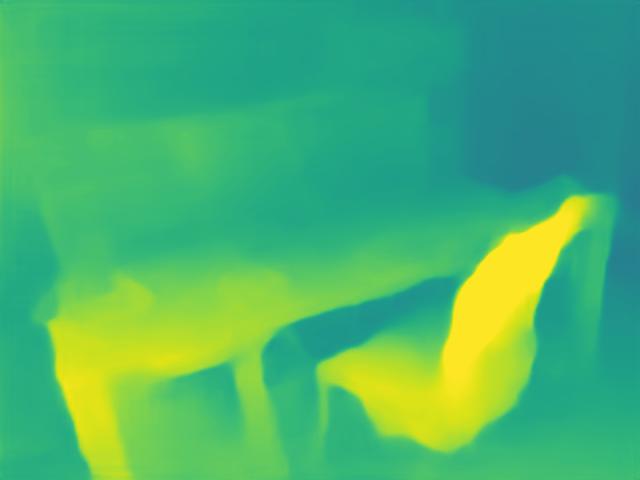}{0}  
  \figcol{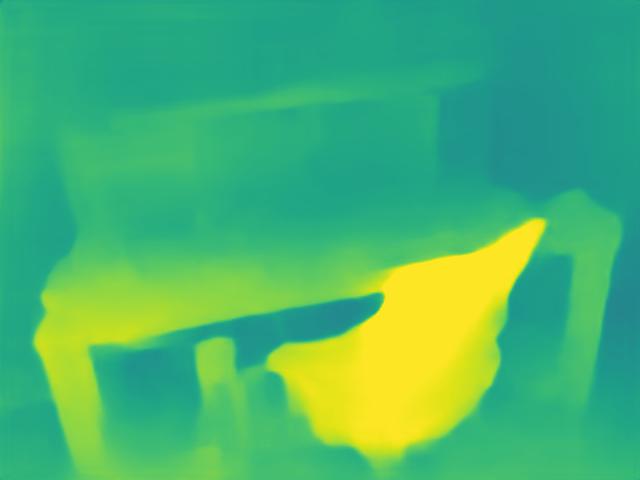}{1.03\figurewidth}
  \figcol{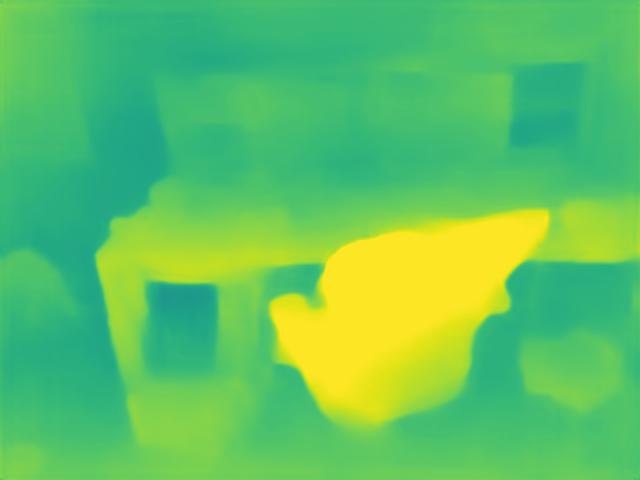}{2.06\figurewidth}
  \figcol{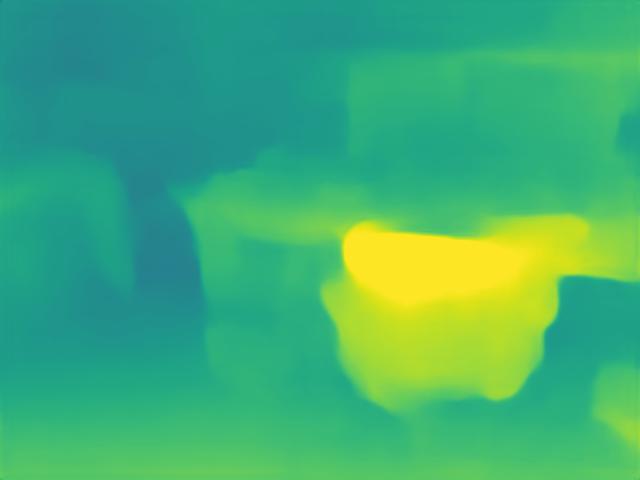}{3.09\figurewidth}
  \figcol{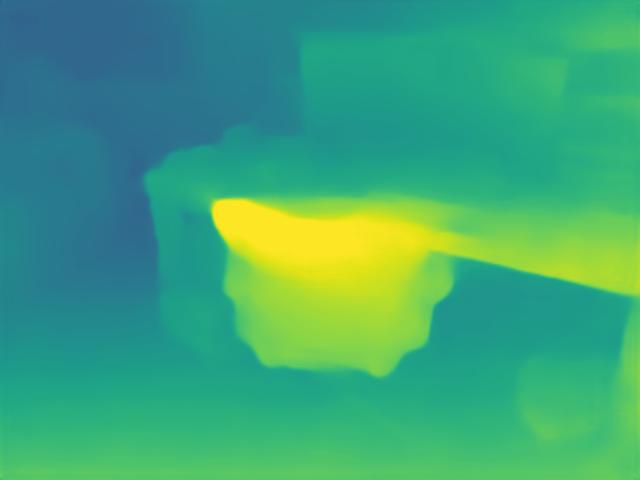}{4.12\figurewidth}

 \draw [draw = red] (-1,1.35) rectangle ++(0.8, 0.6);
  \draw [draw = red] (2.1,1.45) rectangle ++(0.8, 0.6);
    \draw [draw = red] (5.5,1.55) rectangle ++(0.7, 0.7);
   \draw [draw = green] (9.1,1.75) rectangle ++(0.6, 0.7);
   \draw [draw = green] (12,2.05) rectangle ++(0.4, 0.6);
      
  \end{tikzpicture}   
  \caption{Failure cases example. The wrong predictions might be propagated forward because of the fusion in the latent space. However, as the GP only bring a prior for the latent space, the erroneous depth estimates decay away quickly.}
  \label{fig:fail}
\end{figure*}

Fig.~\ref{fig:fail} presents one failure example. As we mentioned, one risk of our method is that wrong predictions can also be propagated forward. In this case, the wrong predictions inside red boxes exist among the first three successive frames, but the erroneous resuts decay away for the latter two frames, as the GP only bring a prior for the latent space and the observations quickly overwhelm it.

\section{Ablation study}
\begin{figure*}[!h]
  \centering
  \setlength{\figurewidth}{.180\textwidth}
  \setlength{\figureheight}{0.75\figurewidth}

  \newcommand{\figg}[1]{\includegraphics[width=.98\figurewidth]{./fig/ablation/redkitchen/#1}}

  \begin{tikzpicture}
        \node[] at  ({0*\figurewidth}, {5.6*\figureheight}) {\scriptsize Frame 15};
        \node[] at  ({1*\figurewidth}, {5.6*\figureheight}) {\scriptsize Frame 16};
        \node[] at  ({2*\figurewidth}, {5.6*\figureheight}) {\scriptsize Frame 39};

      \node[rotate=90] at  ({-0.6*\figurewidth}, {0*\figureheight}) {\scriptsize Mat\'ern};
      \node[rotate=90] at  ({-0.6*\figurewidth}, {1*\figureheight}) {\scriptsize Exponential};
      \node[rotate=90] at  ({-0.6*\figurewidth}, {2*\figureheight}) {\scriptsize TD};
      \node[rotate=90] at  ({-0.6*\figurewidth}, {3*\figureheight}) {\scriptsize w/o GP};
      \node[rotate=90] at  ({-0.6*\figurewidth}, {4*\figureheight}) {\scriptsize Ground-truth};      
      \node[rotate=90] at  ({-0.6*\figurewidth}, {5*\figureheight}) {\scriptsize Reference};    
      
     \node [draw=white,thick,minimum width=\figurewidth,inner sep=0] at
      (0, {0*\figureheight}) {\figg{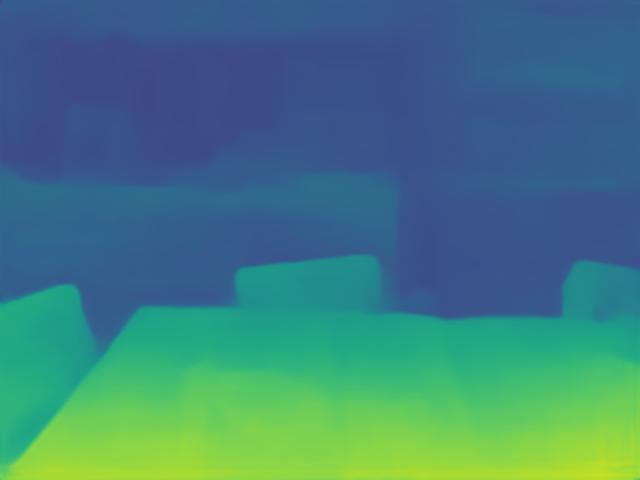}};%
      \node [draw=white,thick,minimum width=\figurewidth,inner sep=0] at
      (0, {1*\figureheight}) {\figg{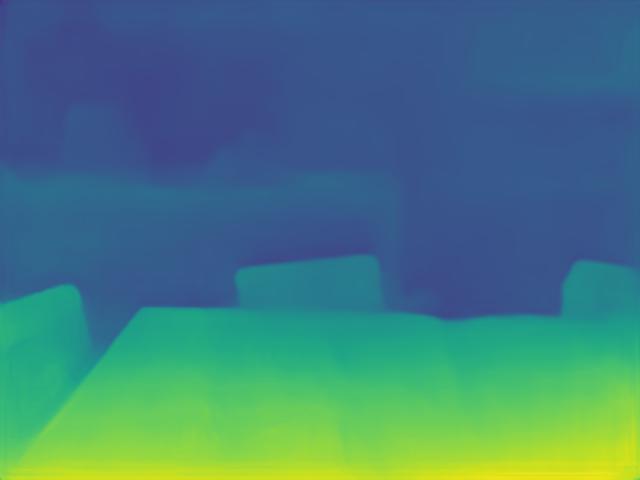}};%
     \node [draw=white,thick,minimum width=\figurewidth,inner sep=0] at
      (0, {2*\figureheight}) {\figg{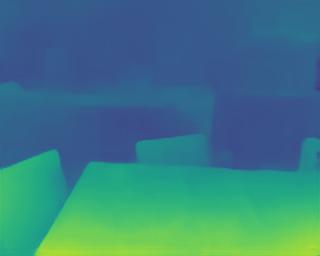}};%
      \node [draw=white,thick,minimum width=\figurewidth,inner sep=0] at
      (0, {3*\figureheight}) {\figg{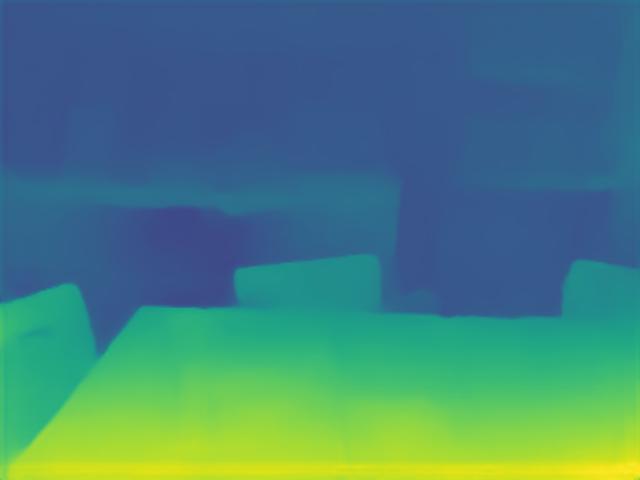}};%
     \node [draw=white,thick,minimum width=\figurewidth,inner sep=0] at
      (0, {4*\figureheight}) {\figg{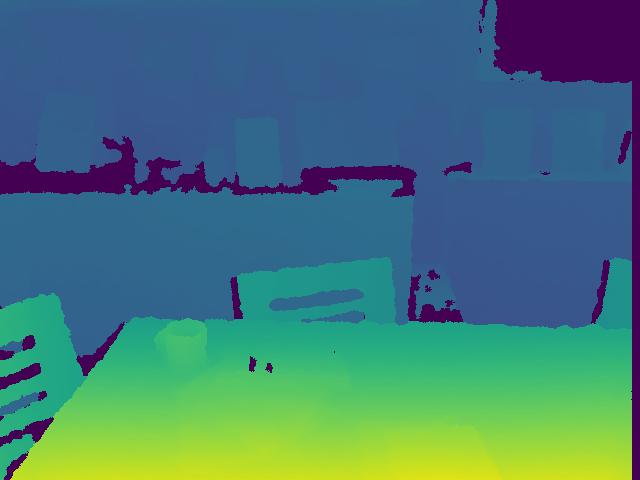}};%
      \node [draw=white,thick,minimum width=\figurewidth,inner sep=0] at
      (0, {5*\figureheight}) {\figg{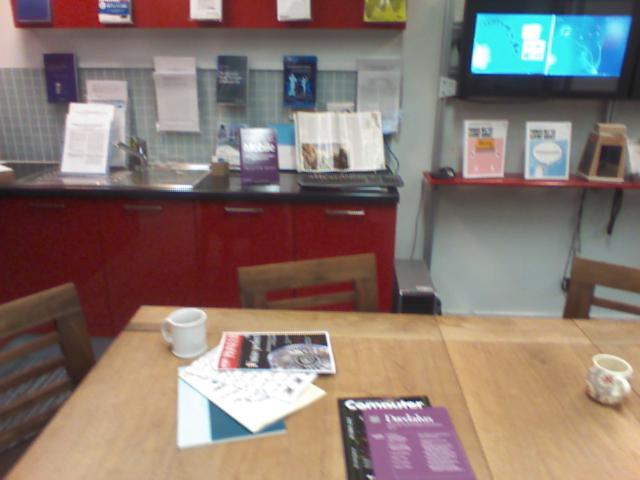}};%
      
      \node [draw=white,thick,minimum width=\figurewidth,inner sep=0] at
      (1.01*\figurewidth, {0*\figureheight}) {\figg{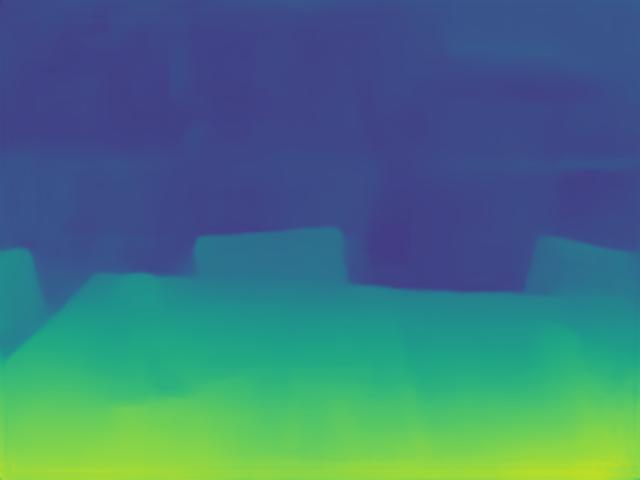}};%
      \node [draw=white,thick,minimum width=\figurewidth,inner sep=0] at
      (1.01*\figurewidth, {1*\figureheight}) {\figg{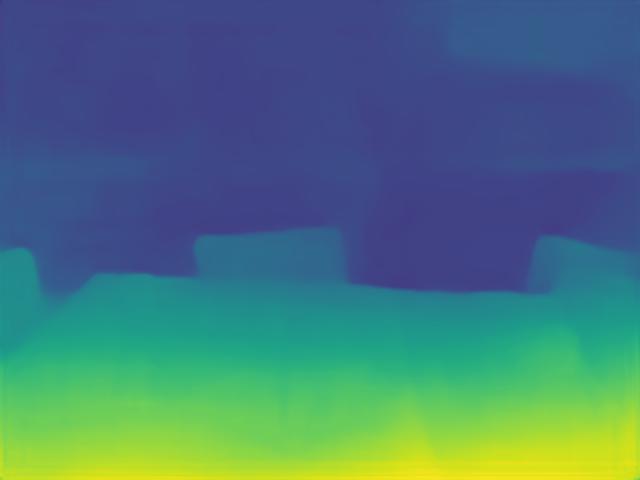}};%
     \node [draw=white,thick,minimum width=\figurewidth,inner sep=0] at
      (1.01*\figurewidth, {2*\figureheight}) {\figg{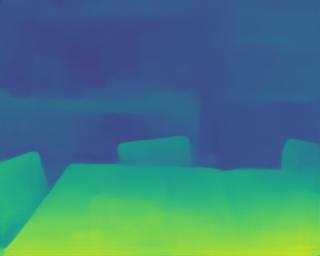}};%
      \node [draw=white,thick,minimum width=\figurewidth,inner sep=0] at
      (1.01*\figurewidth, {3*\figureheight}) {\figg{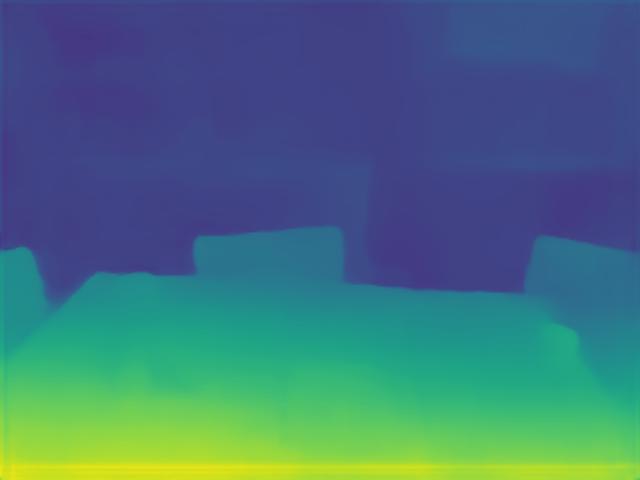}};%
     \node [draw=white,thick,minimum width=\figurewidth,inner sep=0] at
      (1.01*\figurewidth, {4*\figureheight}) {\figg{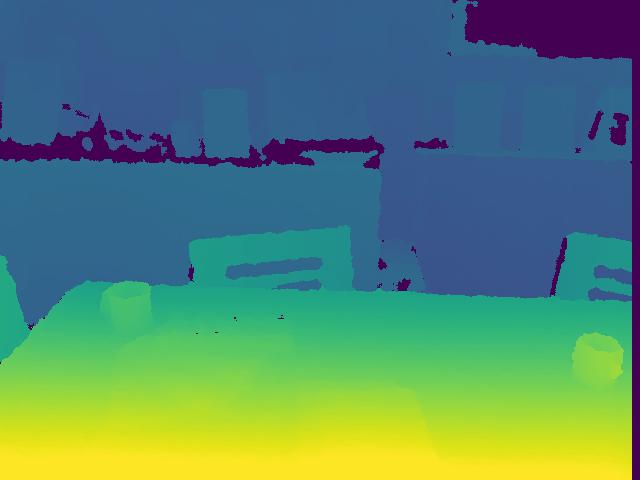}};%
      \node [draw=white,thick,minimum width=\figurewidth,inner sep=0] at
      (1.01*\figurewidth, {5*\figureheight}) {\figg{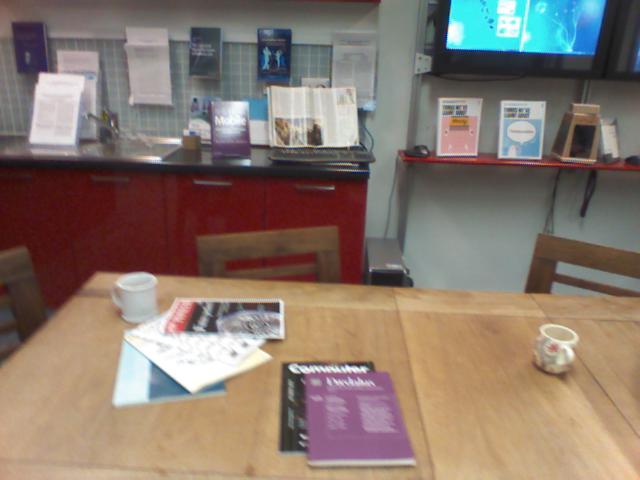}};%
            
      \node [draw=white,thick,minimum width=\figurewidth,inner sep=0] at
      (2.01*\figurewidth, {0*\figureheight}) {\figg{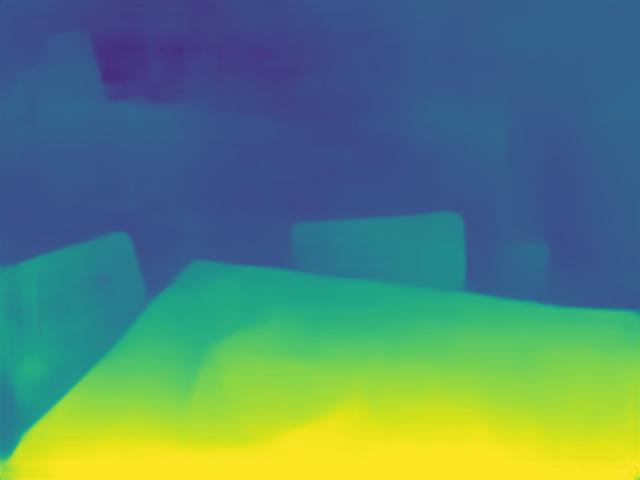}};%
      \node [draw=white,thick,minimum width=\figurewidth,inner sep=0] at
      (2.01*\figurewidth, {1*\figureheight}) {\figg{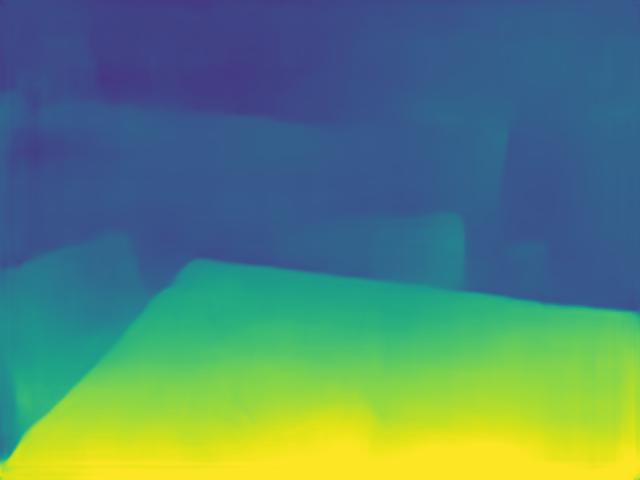}};%
     \node [draw=white,thick,minimum width=\figurewidth,inner sep=0] at
      (2.01*\figurewidth, {2*\figureheight}) {\figg{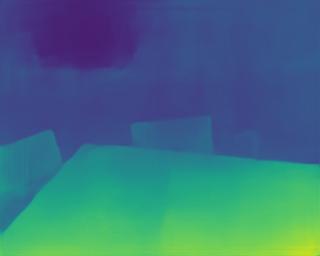}};%
      \node [draw=white,thick,minimum width=\figurewidth,inner sep=0] at
      (2.01*\figurewidth, {3*\figureheight}) {\figg{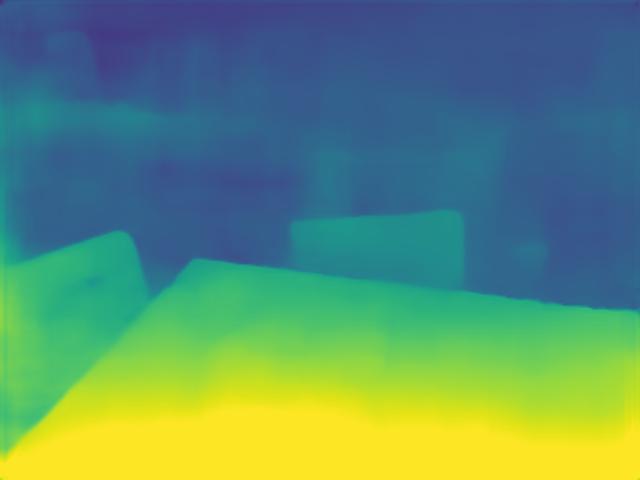}};%
     \node [draw=white,thick,minimum width=\figurewidth,inner sep=0] at
      (2.01*\figurewidth, {4*\figureheight}) {\figg{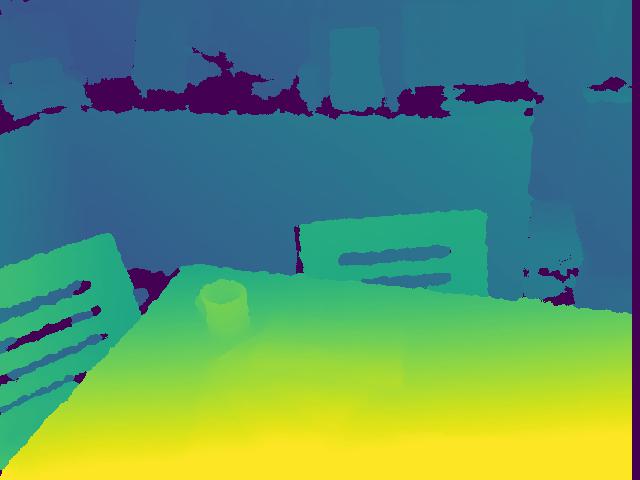}};%
      \node [draw=white,thick,minimum width=\figurewidth,inner sep=0] at
      (2.01*\figurewidth, {5*\figureheight}) {\figg{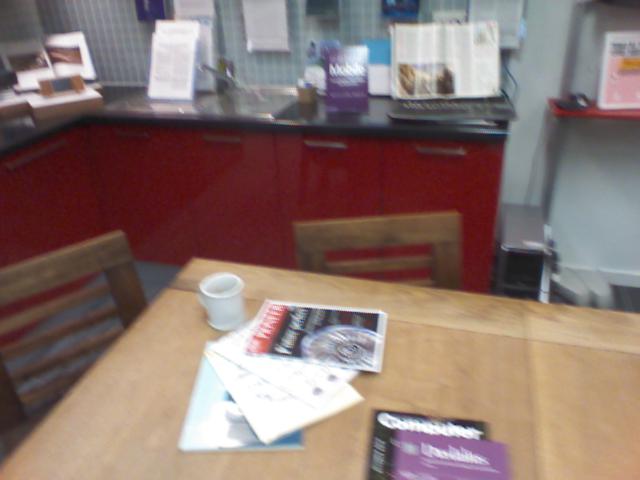}};

      \node [draw=white, minimum width=\figurewidth, inner sep=0] at
      ({3.3*\figurewidth}, {0.5*\figureheight}) {\includegraphics[width=1.46\figurewidth]{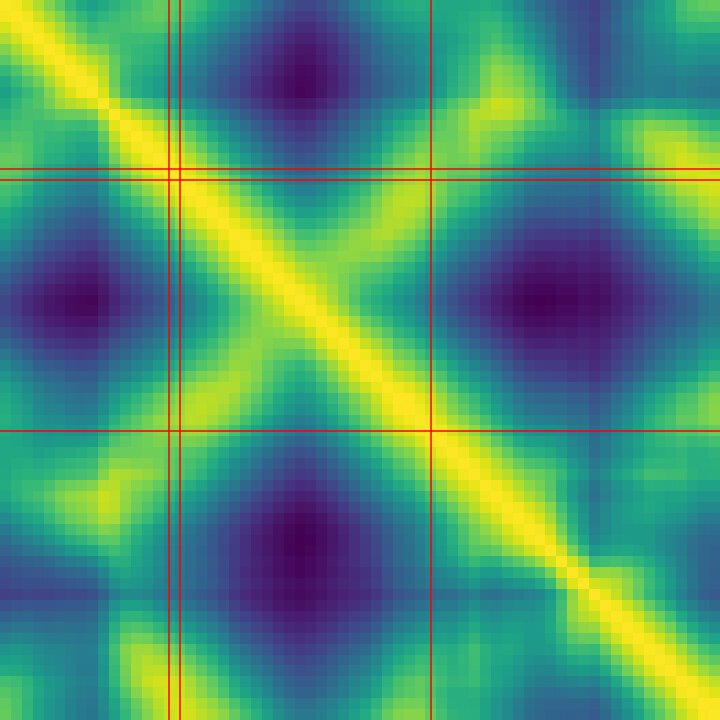}};%
      
      \node[rotate=270] at  ({4.1*\figurewidth}, {0.5*\figureheight}) {\scriptsize Mat\'ern kernel};
      
      \node [draw=white,minimum width=\figurewidth, inner sep=0] at
      ({3.3*\figurewidth}, 2.5*\figureheight) {\includegraphics[width=1.46\figurewidth]{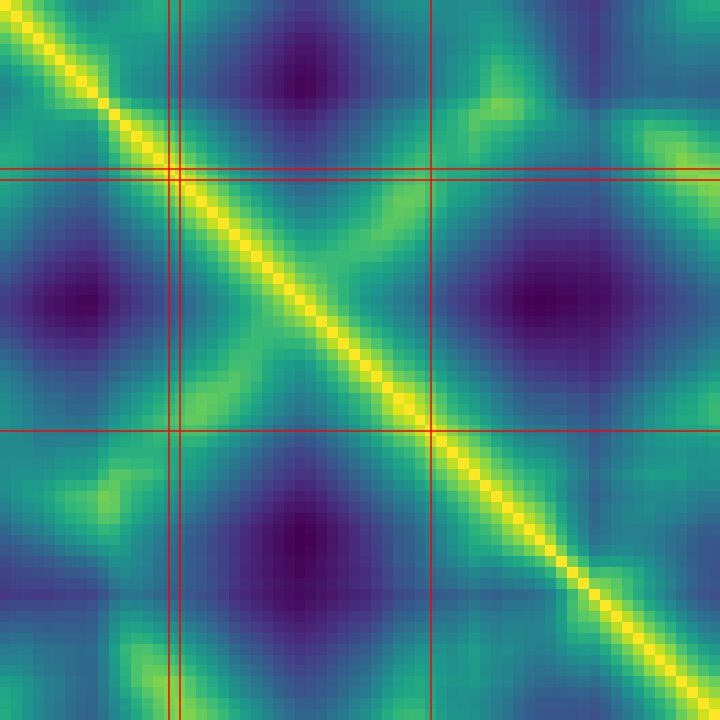}};
      
      \node[rotate=270] at  ({4.1*\figurewidth}, {2.5*\figureheight}) {\scriptsize Exponential kernel};
      
      \node [draw=white, minimum width=\figurewidth, inner sep=0] at
      ({3.3*\figurewidth}, 4.5*\figureheight) {\includegraphics[width=1.46\figurewidth]{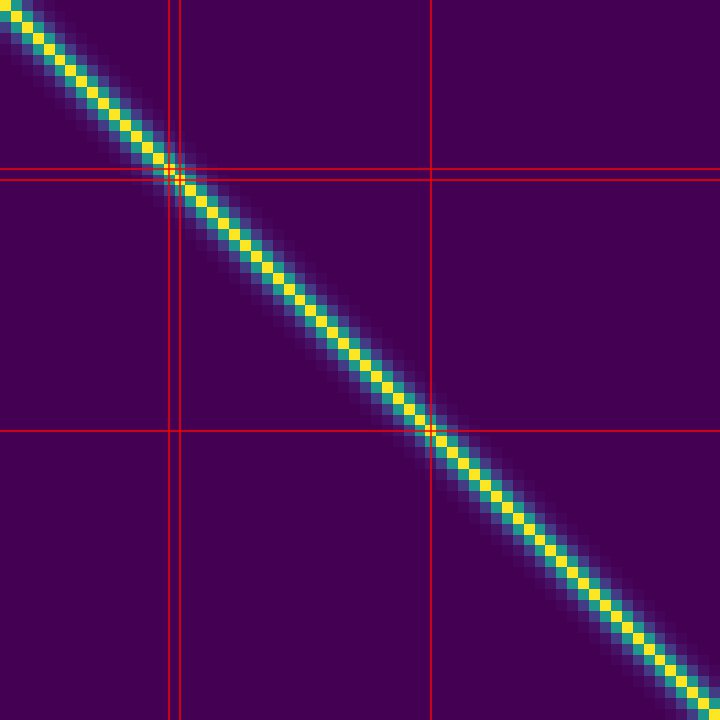}};%
      
      \node[rotate=270] at  ({4.1*\figurewidth}, {4.5*\figureheight}) {\scriptsize TD kernel};

  \end{tikzpicture}
  \caption{Results comparison of different choices of kernel function on the \textit{redkitchen} sequence in \textsc{7scenes}.}
  \label{fig:k1}
\end{figure*}

\begin{figure*}[!h]
  \centering
  \setlength{\figurewidth}{.180\textwidth}
  \setlength{\figureheight}{0.75\figurewidth}

  \newcommand{\figg}[1]{\includegraphics[width=.98\figurewidth]{./fig/ablation/sun3d/#1}}

  \begin{tikzpicture}
  
        \node[] at  ({0*\figurewidth}, {5.6*\figureheight}) {\scriptsize Frame 9};
        \node[] at  ({1*\figurewidth}, {5.6*\figureheight}) {\scriptsize Frame 10};
        \node[] at  ({2*\figurewidth}, {5.6*\figureheight}) {\scriptsize Frame 160};  
 
      \node[rotate=90] at  ({-0.6*\figurewidth}, {0*\figureheight}) {\scriptsize Mat\'ern};
      \node[rotate=90] at  ({-0.6*\figurewidth}, {1*\figureheight}) {\scriptsize Exponential};
      \node[rotate=90] at  ({-0.6*\figurewidth}, {2*\figureheight}) {\scriptsize TD};
      \node[rotate=90] at  ({-0.6*\figurewidth}, {3*\figureheight}) {\scriptsize w/o GP};
      \node[rotate=90] at  ({-0.6*\figurewidth}, {4*\figureheight}) {\scriptsize Ground-truth};      
      \node[rotate=90] at  ({-0.6*\figurewidth}, {5*\figureheight}) {\scriptsize Reference};    
      
     \node [draw=white,thick,minimum width=\figurewidth,inner sep=0] at
      (0, {0*\figureheight}) {\figg{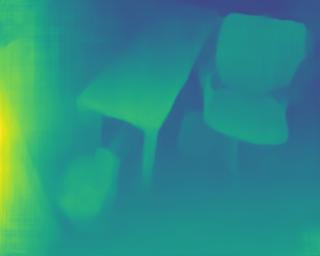}};%
      \node [draw=white,thick,minimum width=\figurewidth,inner sep=0] at
      (0, {1*\figureheight}) {\figg{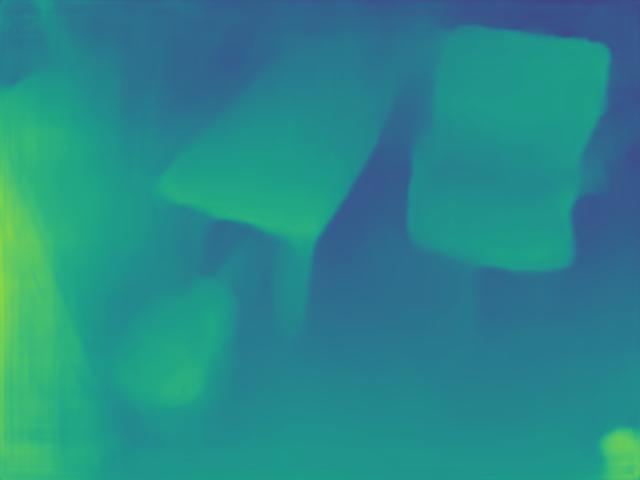}};%
     \node [draw=white,thick,minimum width=\figurewidth,inner sep=0] at
      (0, {2*\figureheight}) {\figg{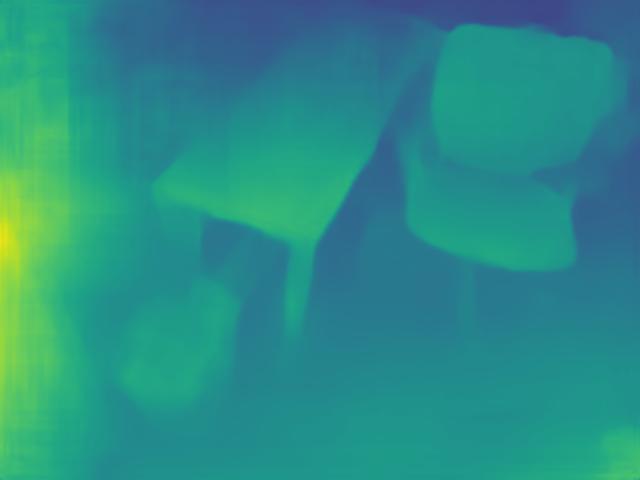}};%
      \node [draw=white,thick,minimum width=\figurewidth,inner sep=0] at
      (0, {3*\figureheight}) {\figg{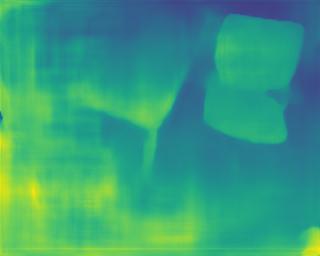}};%
     \node [draw=white,thick,minimum width=\figurewidth,inner sep=0] at
      (0, {4*\figureheight}) {\figg{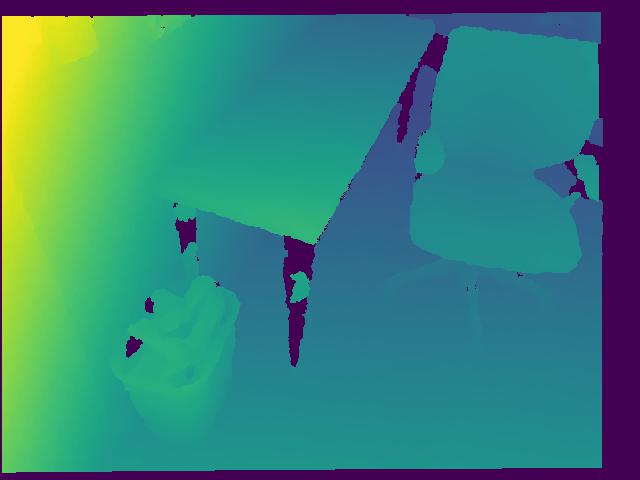}};%
      \node [draw=white,thick,minimum width=\figurewidth,inner sep=0] at
      (0, {5*\figureheight}) {\figg{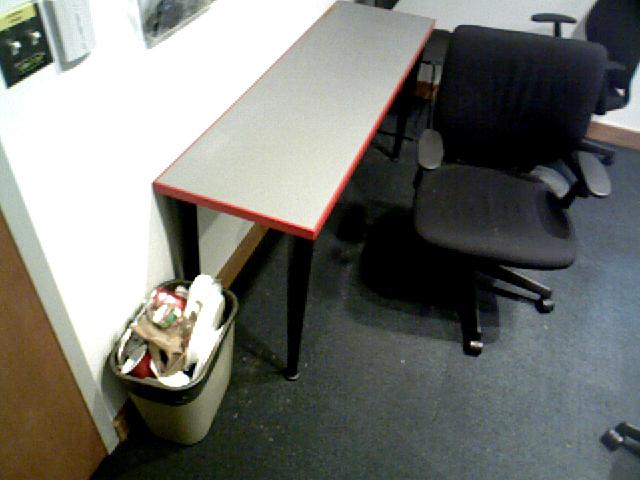}};%
      
      \node [draw=white,thick,minimum width=\figurewidth,inner sep=0] at
      (1.01*\figurewidth, {0*\figureheight}) {\figg{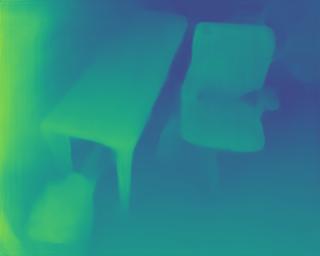}};%
      \node [draw=white,thick,minimum width=\figurewidth,inner sep=0] at
      (1.01*\figurewidth, {1*\figureheight}) {\figg{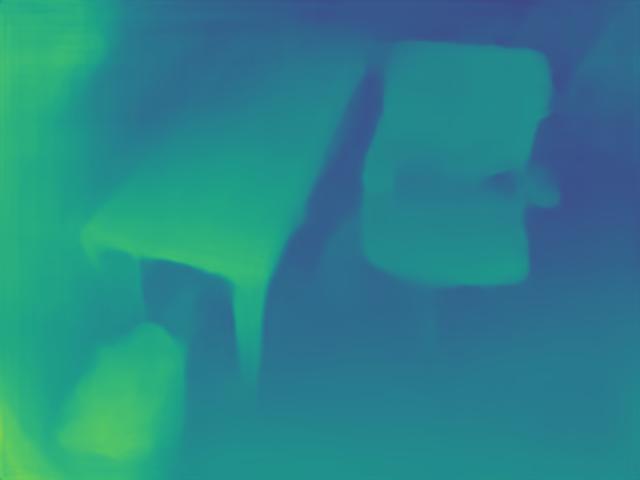}};%
     \node [draw=white,thick,minimum width=\figurewidth,inner sep=0] at
      (1.01*\figurewidth, {2*\figureheight}) {\figg{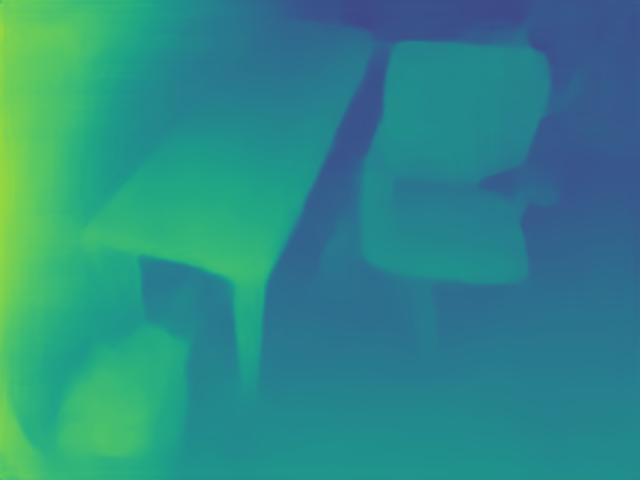}};%
      \node [draw=white,thick,minimum width=\figurewidth,inner sep=0] at
      (1.01*\figurewidth, {3*\figureheight}) {\figg{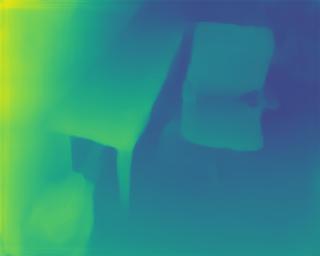}};%
     \node [draw=white,thick,minimum width=\figurewidth,inner sep=0] at
      (1.01*\figurewidth, {4*\figureheight}) {\figg{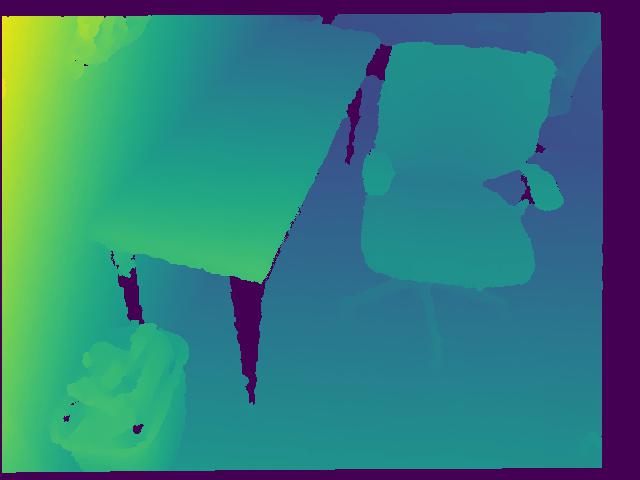}};%
      \node [draw=white,thick,minimum width=\figurewidth,inner sep=0] at
      (1.01*\figurewidth, {5*\figureheight}) {\figg{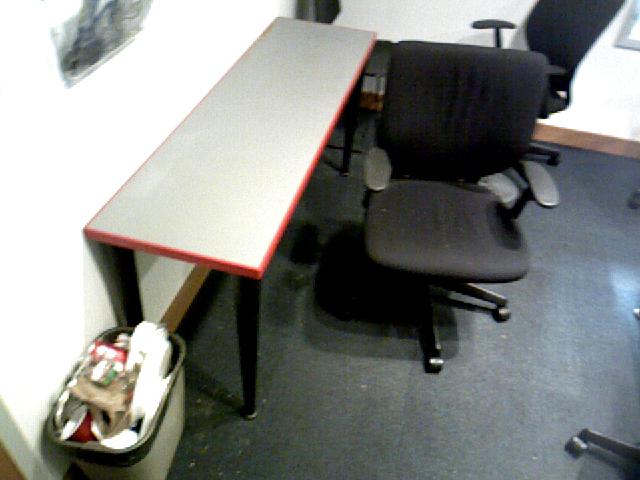}};%
            
      \node [draw=white,thick,minimum width=\figurewidth,inner sep=0] at
      (2.01*\figurewidth, {0*\figureheight}) {\figg{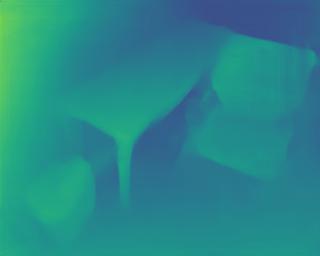}};%
      \node [draw=white,thick,minimum width=\figurewidth,inner sep=0] at
      (2.01*\figurewidth, {1*\figureheight}) {\figg{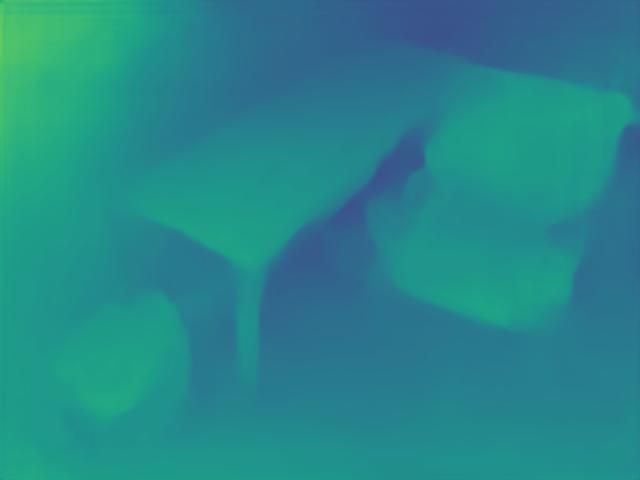}};%
     \node [draw=white,thick,minimum width=\figurewidth,inner sep=0] at
      (2.01*\figurewidth, {2*\figureheight}) {\figg{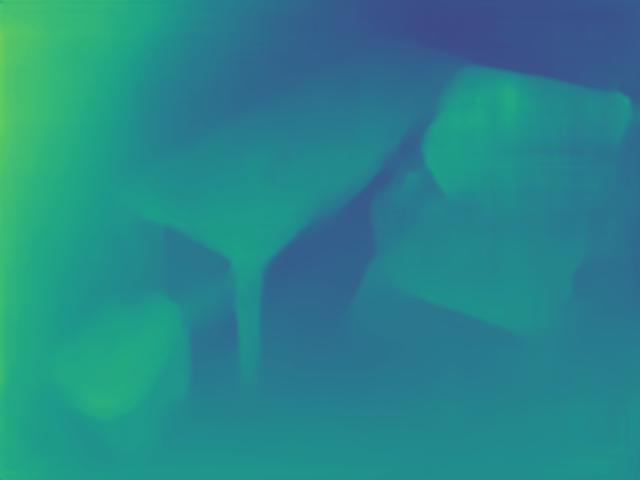}};%
      \node [draw=white,thick,minimum width=\figurewidth,inner sep=0] at
      (2.01*\figurewidth, {3*\figureheight}) {\figg{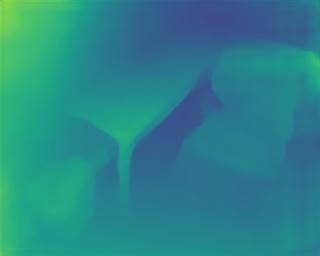}};%
     \node [draw=white,thick,minimum width=\figurewidth,inner sep=0] at
      (2.01*\figurewidth, {4*\figureheight}) {\figg{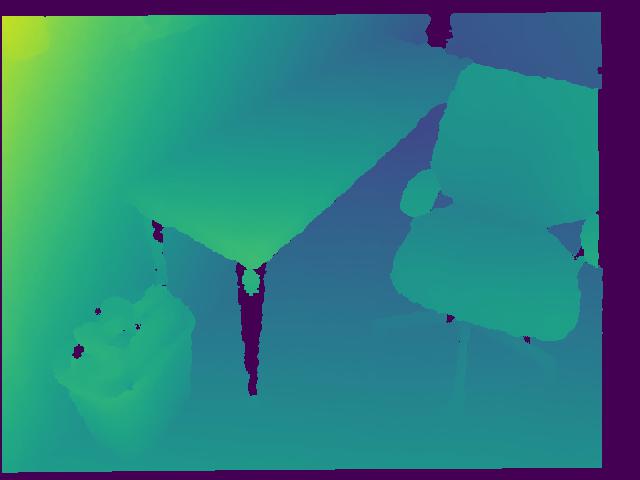}};%
      \node [draw=white,thick,minimum width=\figurewidth,inner sep=0] (foo) at
      (2.01*\figurewidth, {5*\figureheight}) {\figg{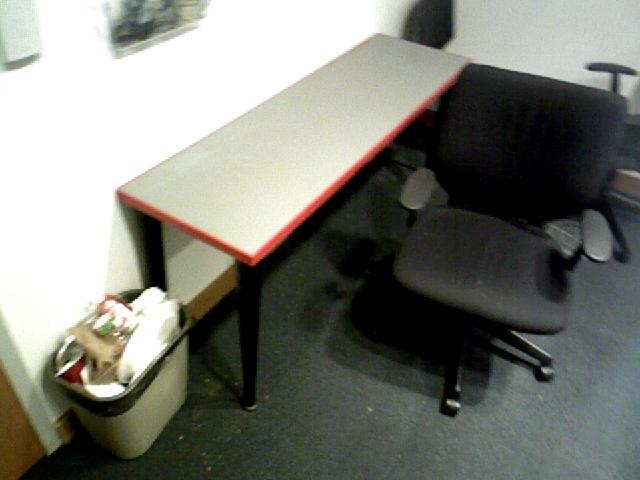}};
            
      \node [draw=white, minimum width=\figurewidth, inner sep=0] at
      ({3.3*\figurewidth}, {0.5*\figureheight}) {\includegraphics[width=1.46\figurewidth]{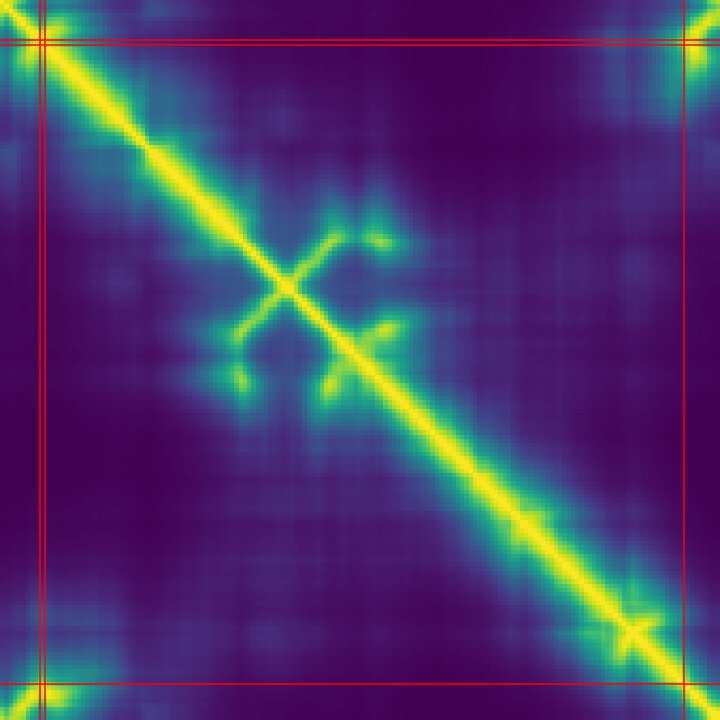}};%
      
      \node[rotate=270] at  ({4.1*\figurewidth}, {0.5*\figureheight}) {\scriptsize Mat\'ern kernel};
      
      \node [draw=white,minimum width=\figurewidth, inner sep=0] at
      ({3.3*\figurewidth}, 2.5*\figureheight) {\includegraphics[width=1.46\figurewidth]{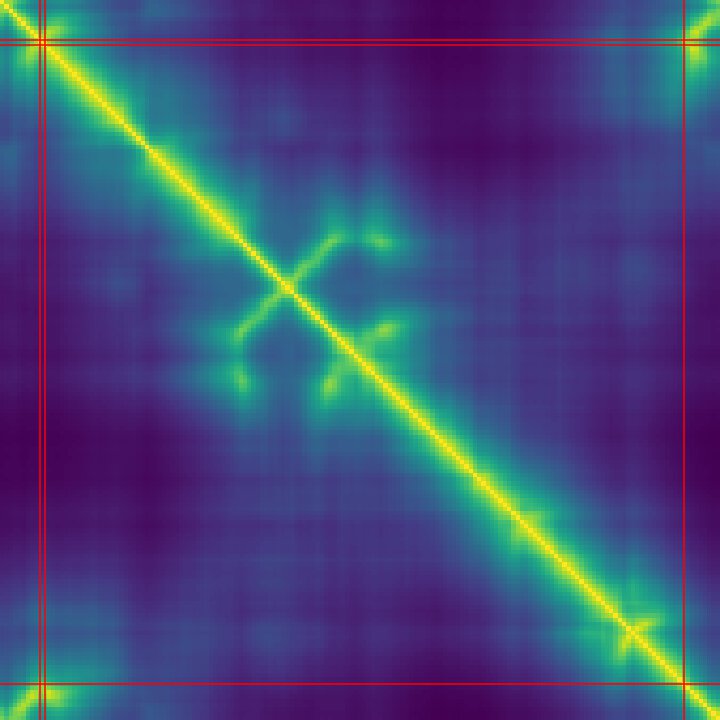}};
      
      \node[rotate=270] at  ({4.1*\figurewidth}, {2.5*\figureheight}) {\scriptsize Exponential kernel};
      
      \node [draw=white, minimum width=\figurewidth, inner sep=0] at
      ({3.3*\figurewidth}, 4.5*\figureheight) {\includegraphics[width=1.46\figurewidth]{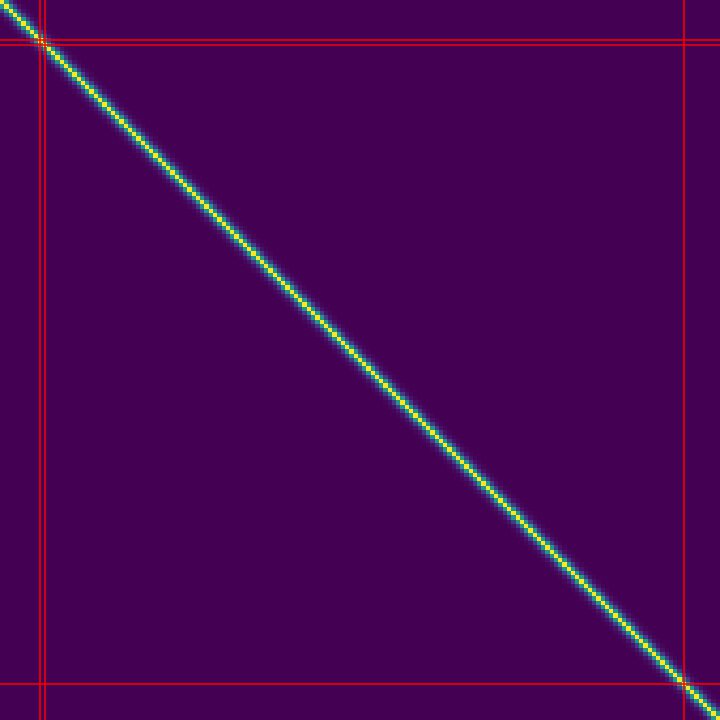}};%
      
      \node[rotate=270] at  ({4.1*\figurewidth}, {4.5*\figureheight}) {\scriptsize TD kernel};

  \end{tikzpicture}
  \caption{Results comparison of different choices of kernel function on \textsc{sun3d}.}
  \label{fig:k2}
\end{figure*}

\noindent
In Sec.~\ref{sec:ablation}, we presented several ablation studies. Here we provide additional qualitative comparisons (to supplement the metrics in the main paper) of different choices of kernel function in Fig.~\ref{fig:k1} and Fig.~\ref{fig:k2}. We visualize the TD kernel, exponential kernel, and Mat\'ern kernel. The results show that the TD kernel is limited to considering the consecutive two neighbour frames as it uses a different distance metric. Additionally, the Mat\'ern kernel has stronger coupling than exponential kernel. For each example, we show results of three frames, including both near neighbour and far neighbour. We use red lines to label the selected frames in the kernel images. It shows that for the far neighbour (see frame 39 in Fig.~\ref{fig:k1} and frame 160 in Fig.~\ref{fig:k2}), the results of the TD kernel and w/o GP are worse than the results of exponential kernel and Mat\'ern kernel, as they cannot leverage information from distant past frames though they share similar views. Comparing the exponential kernel and the Mat\'ern kernel, the results of Mat\'ern have sharper edges.

\section{Supplementary video}
\noindent
The project page (\url{https://aaltoml.github.io/GP-MVS}) features a supplementary video with example sequences from \textsc{7scenes} (\textit{office-04}) and \textsc{sun3d} (\textit{mit\_46\_6lounge}). The benefits of the GP model are apparent especially in the cases where the camera stays still. Furthermore, we have included two example sequences captured from our iPad implementation, where the inference runs in real-time on the device. 

Note that there are no view selection heuristics and we only need to store the previous frame. The effect of the GP can be seen clearly when the app starts up and the GP first accumulates information over frames.

\section{Inference time}
\noindent
We also evaluated the inference time on our desktop mentioned in Sec.~\ref{sec:experiments}: Our method (online) $0.076 {\pm} 0.003$~s, MVDepthNet $0.066 \pm 0.008$~s, DeepMVS $4.9 {\pm} 0.1$~s, MVSNet $3.2 {\pm} 0.1$~s, and COLMAP $4.5 {\pm} 0.5$~s. As we discussed in Sec.~\ref{sec:eval}, these results proves that the improvement introduced by GP come at almost no additional cost.

\end{document}